\documentclass[sigconf]{acmart}

\AtBeginDocument{%
  }

\setcopyright{acmlicensed}
\copyrightyear{2018}
\acmYear{2018}
\acmDOI{XXXXXXX.XXXXXXX}

\acmConference[Preprint]{Preprint}{October 12, 2025}{}

\acmISBN{978-1-4503-XXXX-X/18/06}

\usepackage{graphicx} 
\usepackage{caption} 
\usepackage{subcaption} 
\usepackage{multirow} 
\renewcommand\footnotetextcopyrightpermission[1]{}
\begin{document}

\title{Mental{GLM} Series: Explainable Large Language Models for Mental Health Analysis on Chinese Social Media}

\author{Wei Zhai}
\affiliation{%
  \institution{Beijing University of Technology}
  \city{Beijing}
  \country{China}}
\email{zhaiwei11@emails.bjut.edu.cn}

\author{Nan Bai}
\affiliation{%
  \institution{Wuhan University}
  \city{Wuhan}
  \country{China}}
\email{2024103070005@whu.edu.cn}

\author{Qing Zhao}
\affiliation{%
  \institution{Beijing University of Technology}
  \city{Beijing}
  \country{China}}
\email{zhaoqing@bjut.edu.cn}

\author{Jianqiang Li}
\affiliation{%
  \institution{Beijing University of Technology}
  \city{Beijing}
  \country{China}}
\email{lijianqiang@bjut.edu.cn}

\author{Fan Wang}
\affiliation{%
  \institution{Wuhan University}
  \city{Wuhan}
  \country{China}}
\email{2022203070010@whu.edu.cn}

\author{Hongzhi Qi}
\affiliation{%
  \institution{Beijing University of Technology}
  \city{Beijing}
  \country{China}}
\email{qhz123@emails.bjut.edu.cn}

\author{Meng Jiang}
\affiliation{%
  \institution{Beijing University of Technology}
  \city{Beijing}
  \country{China}}
\email{jiangmeng@emails.bjut.edu.cn}


\author{Xiaoqin Wang}
\affiliation{%
  \institution{Wuhan University}
  \city{Wuhan}
  \country{China}}
\email{xiaoqin_wang78@163.com}

\author{Bing Xiang Yang}
\authornote{Corresponding authors.}
\affiliation{%
  \institution{Wuhan University}
  \city{Wuhan}
  \country{China}}
\email{yangbx@whu.edu.cn}

\author{Guanghui Fu}
\authornotemark[1]
\affiliation{%
  \institution{Sorbonne Université}
  \city{Paris}
  \country{France}}
\email{guanghui.fu@icm-institute.org}

\renewcommand{\shortauthors}{Zhai et al.}

\begin{abstract}
As the prevalence of mental health challenges, social media has emerged as a key platform for individuals to express their emotions.
Deep learning tends to be a promising solution for analyzing mental health on social media. However, black box models are often inflexible when switching between tasks, and their results typically lack explanations. With the rise of large language models (LLMs), their flexibility has introduced new approaches to the field. 
Also due to the generative nature, they can be prompted to explain decision-making processes. However, their performance on complex psychological analysis still lags behind deep learning. In this paper, we introduce the first multi-task Chinese Social Media Interpretable Mental Health Instructions (C-IMHI) dataset, consisting of 9K samples, which has been quality-controlled and manually validated. We also propose MentalGLM series models, the first open-source LLMs designed for explainable mental health analysis targeting Chinese social media, trained on a corpus of 50K instructions. The proposed models were evaluated on three downstream tasks and achieved better or comparable performance compared to deep learning models, generalized LLMs, and task fine-tuned LLMs. We validated a portion of the generated decision explanations with experts, showing promising results. We also evaluated the proposed models on a clinical dataset, where they outperformed other LLMs, indicating their potential applicability in the clinical field. Our models show strong performance, validated across tasks and perspectives. The decision explanations enhance usability and facilitate better understanding and practical application of the models. Both the constructed dataset and the models are publicly available via: \url{https://github.com/zwzzzQAQ/MentalGLM}.
\end{abstract}




\maketitle

\section{Introduction}
In modern society, the rising incidence of mental illness has made concern for mental health a widespread issue~\cite{world2022world}. The World Health Organization (WHO) reports that approximately 3.8\% of the global population suffers from depression~\cite{world2023depressive}, while the prevalence of depression in China is as high as 6.9\%~\cite{huang2019prevalence}. 
People often neglect to manage their emotions or, due to the stigma of mental illness, rarely seek psychological support through medical channels~\cite{yu2020coping}. 
On platforms like X and Weibo, comments under depression-related topics often express negative emotions and mention suicidal thoughts~\cite{cao2019latent}. These trends highlight the need for psychological analysis tools for early detection of mental health issues through social media, enabling timely interventions~\cite{coppersmith2018natural}.

Deep learning has been proven to be an effective solution for language processing, particularly with pre-trained language models (PLMs), like MentalBERT~\cite{ji2021mentalbert} and Chinese MentalBERT~\cite{zhai-etal-2024-chinese} which are specifically designed for social media mental health analysis tasks, have demonstrated strong performance. However, the black-box nature of deep learning models limits their use in mental health analysis because they lack transparency in their decision-making processes ~\cite{sheu2020illuminating}. Additionally, they lack flexibility, as they typically require expensive data annotation and task-specific training for each application. 

Recently, the development of large language models (LLMs) has gained attention in the mental health domain~\cite{he2023towards,demszky2023using}. LLMs are highly flexible due to their ability to handle multiple tasks through user prompts, thanks to their training on diverse datasets~\cite{brown2020language}. Yang et al.~\cite{yang-etal-2023-towards} highlighted LLMs' ability to provide explanations for their decisions, underlining their potential for explainable mental health analysis. However, a considerable performance gap remains between LLMs and deep learning models for mental health tasks, as demonstrated by Qi et al.~\cite{qi2023supervised} and Yang et al.~\cite{yang-etal-2023-towards}. Xu et al.~\cite{xu2024mental} showed that fine-tuning LLMs on varied datasets can substantially boost their performance across multiple mental health tasks. To support LLM fine-tuning for mental health analysis on English social media, Yang et al.~\cite{yang2024mentallama} recently developed the multi-task, multi-source Interpretable Mental Health Instructions (IMHI) dataset. In contrast, within the Chinese community, there is no publicly available dataset for fine-tuning LLMs on mental health analysis instructions, nor is there an open-source LLM dedicated to this purpose. The existing dataset, such as the one proposed by Chen et al.~\cite{chen2023soulchat}, which introduced LLMs for psychological counseling using a dataset of single-round and multi-round empathetic dialogues, is valuable but not sufficient to support comprehensive mental health analysis tasks.

To address these gaps, we constructed the first multi-task Chinese Social Media Interpretable Mental Health Instructions (C-IMHI) dataset, with 9K samples for LLM fine-tuning and evaluation. The dataset was created using a teacher-student architecture with GPT-4 to generate reasoning and explanations based on expert-written examples. We validated its quality through both automated methods and expert evaluation, and manually corrected any incorrect samples to create a high-quality dataset. 
We also developed MentalGLM series, the first Chinese open-source explainable LLMs for mental health analysis, fine-tuned through two steps. MentalGLM achieved better or comparable performance compared to deep learning models and fine-tuned LLMs across three downstream tasks, while also providing explainable predictions.  Expert evaluation showed the explanations had high consistency and reliability, comparable to GPT-4. The model's generalization was also tested on cognitive pathway extraction using clinical data, achieving better prediction accuracy than other LLMs, demonstrating potential for clinical applications.

Our contributions are summarized as follows: 1) We have constructed the first dataset for interpretable mental health analysis instructions fine-tuning for Chinese social media, termed C-IMHI, which can also serve as an evaluation benchmark; 2) We introduce MentalGLM series, the first open-source LLMs dedicated to explainable mental health analysis on Chinese social media which surpass or are comparable to SOTA methods across three mental health tasks while providing credible explanations.

\section{Methods} \label{sec:methods}

\begin{figure}[!ht]
  \centering
  \includegraphics[width=0.9\linewidth]{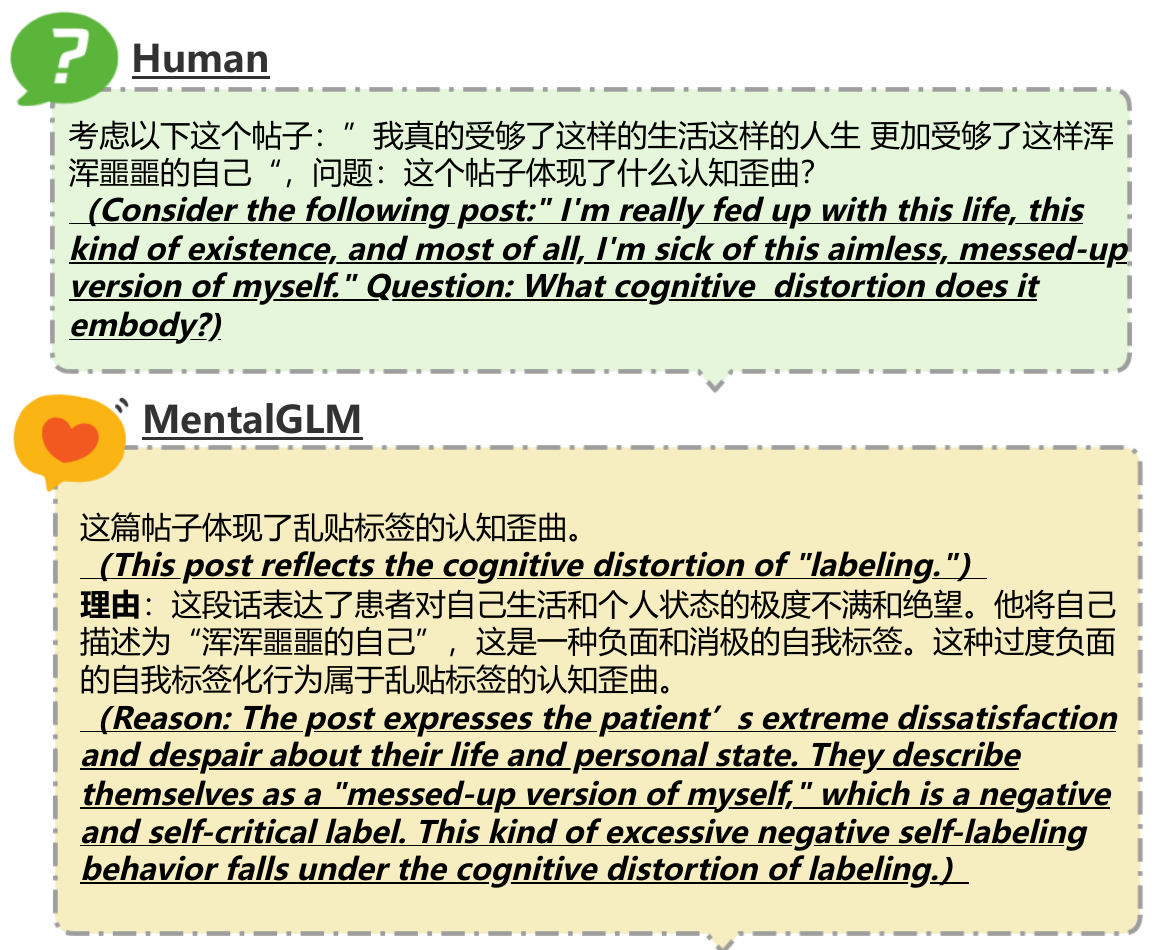}
  \caption{Example of MentalGLM's prediction in the cognitive distortion classification task, including both the prediction and the explanation of its decision.}
  \label{fig:model_output}
\end{figure}

We proposed the MentalGLM series, fine-tuned from the open-source LLMs~\cite{glm2024chatglm}, for mental health analysis on Chinese social media. The two-stage fine-tuning involved first using translated general mental health data (IMHI), then refining with the proposed Chinese-specific social media data (C-IMHI), improving accuracy and explainability. An example of model usage is shown in Figure~\ref{fig:model_output}.

\subsection{Task definition} 
We frame the mental health analysis task as a generative problem using a generative model, specifically an auto-regressive language model $P\phi (O|I)$ with pre-trained weights $\phi$ which generates output $O$ based on input $I$. 
Unlike traditional deep learning models, which require task-specific architectures due to fixed input and output formats and lack explainability, our approach allows for flexible, generative modeling. This enables the simultaneous training of $n$ mental health-related tasks by providing interpretable instructions, allowing the model to generate explanations alongside its predictions. 
Let the dataset $D = \{(I_i, O_i) \}_{i=1}^n $consist of $n$ tasks, where the $i$-th task is represented as a pair of input $I_i$ and output response $O_i$. The input $I_i$ for $i$-th task consists of three components: the task description $d_i$, the text to be processed $t_i$, and the task execution query $q_i$ related to the task. Thus, $I_i = (d_i, t_i, q_i)$. The output response $O_i$ consists of two elements: the required outcome $c_i$ (such as a rating score or classification categories) and the explanation of the decision-making process $e_i$. Therefore, $O_i = (c_i, e_i)$. In summary, each task in the dataset can be formally represented as:
\begin{equation}
\label{eq:data}
D = \{(I_i, O_i)\}_{i=1}^n = \{ ((d_i, t_i, q_i), (c_i, e_i)) \}_{i=1}^n
\end{equation}
By leveraging the interpretable instructions dataset described in Equation~\ref{eq:data}, the foundation model learns to reason from input to output, thereby generating explainable results.

\subsection{Model adaptation from general to mental health analysis}
While open-source Chinese LLMs like GLM~\cite{glm2024chatglm} have demonstrated strong performance in general tasks, they struggle with domain-specific tasks such as mental health analysis~\cite{qi2023supervised}. Instruction fine-tuning has proven to be an effective solution for improving performance in these specialized areas while maintaining the flexibility of LLMs~\cite{yang2024zhongjing}.
Research had shown that the instruction data used for fine-tuning must be sufficiently diverse to ensure the generalization and robustness of the model~\cite{wei2022finetuned}, while also maintaining the uniformity and consistency of the responses~\cite{zhou2024lima}.
However, there is a lack of interpretable instruction fine-tuning datasets in Chinese for mental health analysis tasks. To address this, we first translated the IMHI dataset proposed by Yang et al.~\cite{yang2024mentallama}, which contains multi-task English mental health instruction data, into Chinese for use in the initial stage of our fine-tuning process. The IMHI dataset is designed for developing and validating explainable mental health analysis models, sourced from social media. It is formatted using predefined templates to ensure robust consistency for model training as described in Equation~\ref{eq:data}. The details of the dataset can be seen in Section~\ref{sec:data:english}.
We employed the low-rank adaptation (LORA)~\cite{hu2022lora}, a parameter-efficient adaptation method, to train GLM-4-9b and GLM-4-9b-chat on the translated IMHI training set for five epochs. The best model was selected as the starting point for the next stage of fine-tuning, based on the results from the validation set.

\subsection{Model fine-tuning for Chinese data and task specificity}
\begin{figure}[h]
  \centering
  \includegraphics[width=0.9\linewidth]{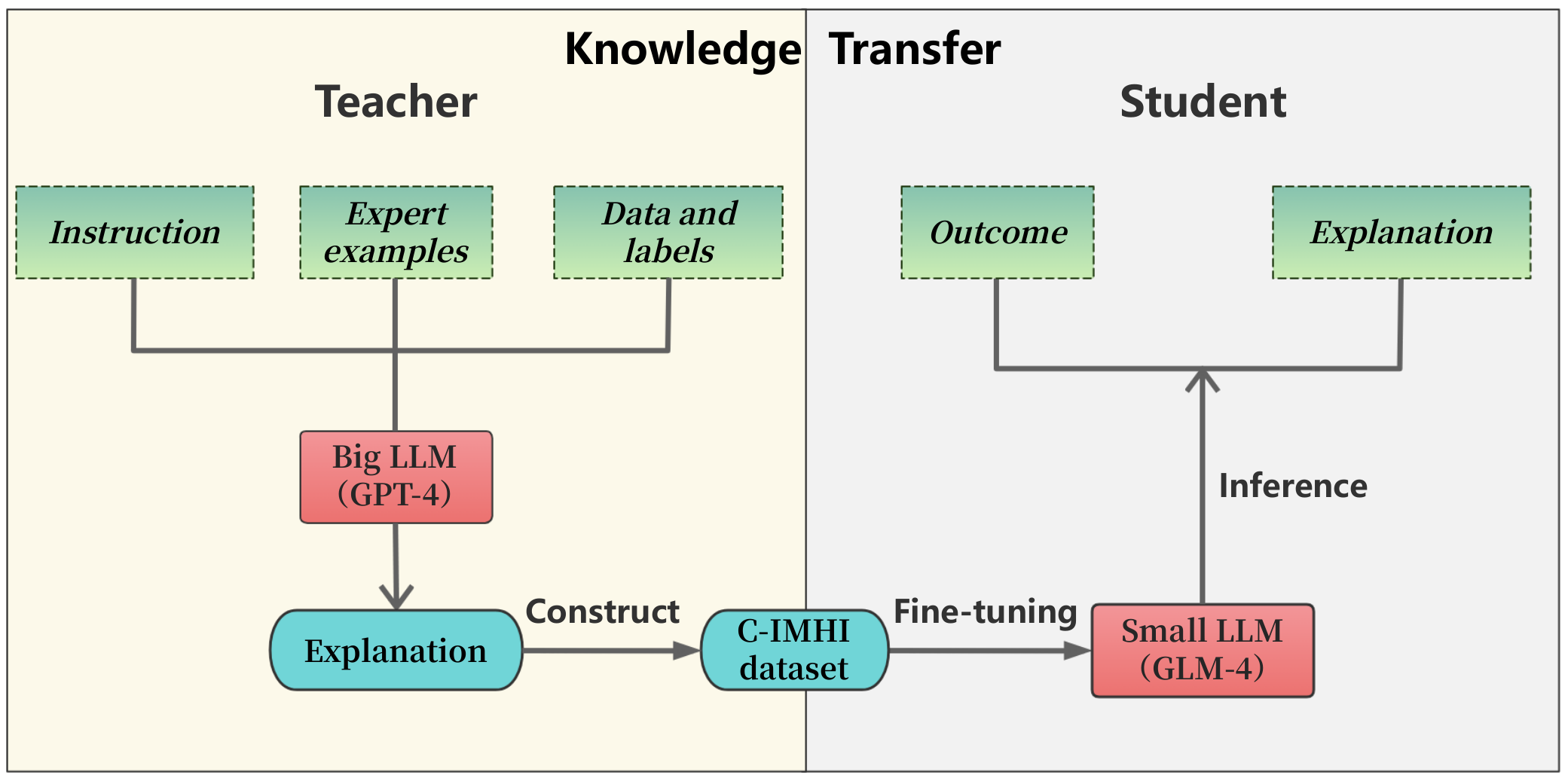}
  \caption{The knowledge transfer process from a larger LLM (GPT-4) to a smaller one (GLM-4-9b), acting as teacher and student respectively.}
  \label{fig:knowledge_transfer}
\end{figure}

The second stage involves adapting the model to the domain of mental health analysis tasks within the specific context of Chinese social media. 
We created a Chinese mental health analysis instruction fine-tuning dataset, named C-IMHI, following the format of the IMHI dataset and including three tasks. 
These three open-source datasets only contain expert annotations without explanations for the decision-making process. To address this, we created the C-IMHI dataset by inviting experts to provide decision-making explanations for a portion of the dataset in each category. We used the advanced LLM GPT-4 to simulate the expert explanation style and generate explanations for the entire dataset.
The idea behind this approach is knowledge transfer, where knowledge from an advanced but expensive model is distilled into a smaller student model to enhance its performance. 
This process is illustrated in Figure~\ref{fig:knowledge_transfer}.
The C-IMHI dataset was evaluated using automated methods, with a subset of samples evaluated by experts.
The experts revised any incorrect samples to ensure the dataset maintains high quality.
Details of this dataset are provided in Section~\ref{sec:data:chinese}, and the evaluation process is described in Section~\ref{sec:exp:dataset_evaluate}. We divided the C-IMHI dataset into a training set, a validation set, and a test set. Building on the best checkpoint from stage~(I), we continued using the LORA method to train on the C-IMHI training set for 10 epochs. The model that showed the best performance on the validation set was used for further evaluation.

\section{Datasets}
The training datasets used in this study consist of two parts: the IMHI dataset, which was used for initial fine-tuning from general to mental health domains, and the proposed C-IMHI dataset, which was used for Chinese-specific fine-tuning and validation. Additionally, we evaluated the trained model on a clinical dataset.

\subsection{English dataset for initial model fine-tuning} \label{sec:data:english}
IMHI is the first multi-task, multi-source interpretable mental health instruction dataset, consisting of 105K data samples, designed to support LLM instruction adaptation and evaluation proposed by Yang et al.~\cite{yang2024mentallama}. 
It includes tasks such as depression detection, stress detection, and mental disorder detection. We utilized a portion of the dataset (some of which is not yet open source) and translated it from English to Chinese. Data distribution is shown in Table~\ref{tab:dataset_general}.
\begin{table*}
  \caption{Data distribution of the IMHI dataset. }
  \label{tab:dataset_general}
  \begin{tabular}{ccccccc}
    \toprule
    Data & Task & Instruction (train/val) & Source & Annotation & Type \\
    \midrule
    DR & depression detection & 1654/184 & Reddit & weak supervision & binary \\
    Dreaddit & stress detection & 3195/356 & Reddit & human annotation & binary \\
    Loneliness & loneliness detection & 477/54 & Reddit & human annotation & binary\\
    SWMH & mental disorders detection & 9793/1089 & Reddit  & weak supervision & multi-class\\
    T-SID & mental disorders detection & 863/96 & Twitter  & weak supervision & multi-class\\
    SAD & stress cause detection & 6162/685 & SMS  & human annotation & multi-class \\
    CAMS & depression/suicide cause detection & 562/63 & Reddit & human annotation & multi-class \\
    MultiWD & wellness dimensions detection & 17716/1969 & Reddit  & human annotation & multi-label \\
    IRF & interpersonal risk factors detection & 6336/705 & Reddit  & human annotation & multi-label  \\
    \midrule
    In total & Mental health analysis & 46758/5201 & Social Media  & - & -\\
    \bottomrule
  \end{tabular}
\end{table*}

We categorize the dataset based on the way of task modeling:
\begin{itemize}
\item \textbf{Binary classification tasks} These tasks aim to determine whether a sample indicates a specific mental health condition, such as depression detection (DR dataset)\cite{pirina2018identifying}, stress detection (Dreaddit dataset)\cite{turcan2019dreaddit}, and loneliness detection (Loneliness dataset)~\cite{yang2024mentallama}.
\item \textbf{Multi-class classification tasks}
These tasks identify the mental health state or underlying cause reflected in a post by selecting the appropriate label from multiple categories. The SWMH~\cite{ji2022suicidal} and T-SID~\cite{ji2022suicidal} datasets detect disorders like suicide risk and depression, while the SAD~\cite{mauriello2021sad} and CAMS~\cite{garg-etal-2022-cams} datasets focus on identifying causes of stress, depression, and suicide, such as work and social relationships.
\item \textbf{Multi-label classification tasks} These tasks assign posts to multiple categories simultaneously. The MultiWD dataset~\cite{sathvik2023multiwd} labels psychological states across dimensions such as psychological, physical, and intellectual. The IRF dataset~\cite{garg-etal-2023-annotated} annotates risk factors for mental disorders.
\end{itemize}

\subsection{Chinese dataset for language and downstream task fine-tuning} \label{sec:data:chinese}
As mentioned earlier, we translated the IMHI dataset from English to Chinese for the model's psychological task fine-tuning. However, the translated text struggles to fully capture the nuances of content within the context of Chinese social media. To address this, we collected three open-source datasets of psychological analysis tasks from Chinese social media (Weibo) for dataset construction as shown below. We invited psychology experts to provide decision making explanations for these representative data. The dataset distribution can be seen in  Table~\ref{tab:dataset_chinese}.
\begin{table*}
  \caption{Data distribution of the proposed C-IMHI dataset.}
  \label{tab:dataset_chinese}
  \begin{tabular}{ccccccc}
    \toprule
    Data & Task & Instruction (train/val/test) & Source & Annotation & Type \\
    \midrule
    SOS-HL-1K & suicide risk detection & 749/250/250 & weibo & human annotation & binary \\
    SocialCD-3K & cognitive distortion detection & 2043/682/682 & weibo & human annotation & multi-label \\
    CP & cognitive path extraction & 2740/910/945 & weibo & human annotation & multi-label\\
    \midrule
    In total & Mental health analysis & 5532/1842/1877 & weibo  & human annotation & -\\
    \bottomrule
  \end{tabular}
\end{table*}

\begin{itemize}
\item \textbf{Suicide risk detection} SOS-HL-1K~\cite{qi2023supervised} is from Weibo, specifically collected from the ``Zoufan'' tree hole\footnote{\url{https://m.weibo.cn/detail/3424883176420210}}. The suicide risk task aims to differentiate between high and low suicide risk. It includes a total of 1,249 posts, and we invited domain experts to provide 22 explanations for representative data—11 for low-risk cases and 11 for high-risk cases.
\item \textbf{Cognitive distortion detection} SocialCD-3K~\cite{qi2023supervised} is from Weibo, also sourced from the ``Zoufan'' tree hole. The cognitive distortion task centers on the categories defined by Burns~\cite{burns1981feeling}. This task is a multi-label classification task, as each post may reflect multiple cognitive distortions across 12 categories. It includes a total of 3,407 posts, and domain experts were invited to provide 28 explanations, with at least two examples for each category.
\item \textbf{Cognitive pathway extraction} CP~\cite{jiang2024ai} is derived from two sources: primarily from Weibo and a smaller portion from Reddit. The data from the Reddit platform were translated into Chinese and subsequently manually checked by annotators. According to the theory of cognitive behavioral therapy (CBT)~\cite{beck1970cognitive}, this task has been defined as a hierarchical text classification task (HTC) with four parent classes and nineteen child classes. Each piece of data may exhibit label classification results of multiple cognitive paths, thus it is categorized as the task category of hierarchical multi-label text classification (HMTC). 
A total of 555 posts were collected and segmented into 4,595 sentences, with experts providing 28 explanations that encompass all four parent classes and nineteen sub-classes.
\end{itemize}

We then used GPT-4 to supplement these explanations for all the data, resulting in the Chinese Social Media Interpretable Mental Health Instruction (C-IMHI) dataset, which contains 9,251 samples. The prompt for explanation generation can be seen in Section~\ref{sec:append:template}.
The use of LLM for generative tasks has demonstrated feasibility~\cite{yu2024large}, especially in the field of mental health, where LLMs have been shown to generate human-level explanations~\cite{yang-etal-2023-towards,yang2024mentallama}.
The prompt of GPT consists of task-specific instructions, data from the original dataset and the corresponding true labels, along with category-specific expert explanations as reference as shown in Figure~\ref{fig:appendix:explain_template}. This process enables GPT to learn and understand the thinking and reasoning processes of human domain experts through examples, while ensuring that the correct answers are provided. 
Furthermore, we designed a comprehensive evaluation process that includes automated methods and a random selection of 100 samples for human evaluation, as detailed in Section~\ref{sec:exp:dataset_evaluate}.
Experts were invited to manually revise any incorrect samples.
We used these data to fine-tune the models and evaluate their performance on downstream tasks. The training and validation sets were used to develop the model, while an independent test set was used to assess and report its performance.

\subsection{Clinical dataset for model evaluation}
We collected cognitive correction materials from 50 patients, comprising 298 sentences, to validate the model's performance on the clinical cognitive pathway extraction task. 
All the patients were diagnosed with mood disorders and demonstrating high treatment compliance. The data was collected between November 11 and November 23, 2023, in the depression ward of Wuhan Mental Health Hospital, China.
The study was approved by the Institutional Ethical Committee (Wuhan University Biomedical Ethics Committee, WHU-LFMD-IRB2023021) and informed consent was obtained from both the hospital and the patients. 
All data were anonymized to protect patient privacy.
Patients were instructed to document their thoughts in a structured format, including: 1) the event that triggered feelings of sadness, anger, or despair, 2) their thoughts about the event, 3) emotional and behavioral responses, and 4) self-refutation to challenge initial thoughts. This method serves as a self-practice tool for cognitive correction and offers psychologists deeper insights into the patient's cognitive patterns. 
However, even with the structured format, patients often struggle to accurately express their events or emotions, highlighting the need for automated tools to extract and analyze cognitive pathways.

\section{Experiments}
We designed experiments to validate both the quality of the C-IMHI dataset and the performance of the proposed MentalGLM series models. The evaluation includes ablation studies, performance assessments on three downstream tasks, and an evaluation of the quality of the model-generated explanations. We also evaluated the trained model on a clinical dataset to assess its generalization capabilities.

\subsection{Implementation details}
GLM-4-9b\footnote{https://huggingface.co/THUDM/glm-4-9b}  is the open source version of the latest generation of pre-trained models launched by Zhipu AI. In the evaluation of datasets in semantics, mathematics, reasoning, code, and knowledge, GLM-4-9b and its human preference-aligned version GLM-4-9b-Chat \footnote{https://huggingface.co/THUDM/glm-4-9b-chat} both showed superior performance beyond Llama-3-8B~\cite{glm2024chatglm}.
We developed the MentalGLM series from GLM-4, resulting in two versions: MentalGLM and MentalGLM-chat, based on GLM-4-9b and GLM-4-9b-chat, respectively.
During the two-step model training process, we set the batch size to 4 and employed a gradient accumulation strategy with 8 accumulation steps, resulting in an effective batch size of 32. Model training utilized the AdamW~\cite{loshchilov2018decoupled} optimizer with a maximum learning rate of 1e-4 and a warm-up ratio of 1\%. Furthermore, the maximum input length of the model was set to 1024. Additionally, to enhance computational efficiency, the float 16 numerical format was employed. All experiments were conducted on a single NVIDIA Tesla V100 32GB SXM2 GPU. 
All the development details, codes and proposed models are publicly available via: 
\url{https://github.com/zwzzzQAQ/MentalGLM}.

\subsection{Quality evaluation of the C-IMHI dataset} \label{sec:exp:dataset_evaluate}
To ensure the quality of the C-IMHI dataset we constructed,we followed the evaluation metrics used to evaluate IHMI~\cite{yang2024mentallama} dataset.
Given the extensive size of the dataset, all generated explanations were evaluated automatically. A subset of 100 samples was randomly selected for detailed human evaluation.

\subsubsection{Automated evaluation} 
In automated evaluation of the C-IMHI dataset, we focused on two key aspects: correctness and consistency: 
1) Correctness: the generated prediction should be correct compare with the ground truth~\cite{yang2024mentallama}; 
2) Consistency: the generated explanations should provide a reasoning process that explains the decision basis and is consistent with the prediction~\cite{wang2023selfconsistency}. 

\textbf{Correctness} In the process of generating explanations, we incorporate annotated samples (data with annotations) and expert-provided examples into the prompt to guide GPT-4 in producing explanations. However, during the experiment, we noted that GPT-4 sometimes contradicted the provided annotations and produced explanations at odds with the established labels. It can be easily filtered using regular expressions for keyword detection, and in cases of incorrect annotations or explanations, we requested experts to revise them. We calculated the ``agreed'' and ``disagreed'' rates for each dataset as the evaluation metrics.

\textbf{Consistency} All GPT-4 generations follow a fixed template, as illustrated in Section~\ref{fig:appendix:explain_template}. Consistency assessment aims to verify whether each explanation supports its respective label. To achieve this, we trained a deep learning model using explanation-label pairs. The same data split used in downstream evaluation was maintained, with generated explanations replacing social media posts. The hypothesis is that if the model achieves high performance, it captures consistent patterns aligning explanations with labels, demonstrating data consistency. If the model performs well on the test set without a significant gap, it further confirms generalization to unseen data. Details of the model training process can be found in Section~\ref{sec:append:quality_eval}. Finally, we applied the trained model to both the test set and the expert-provided example set, using F1-scores as the evaluation metric.

\subsubsection{Human evaluation}
\label{sec:exp:dataset_evaluate:human}
We randomly selected one hundred explanations generated by GPT-4 for subsequent expert evaluation. The evaluation method was based on the scheme used in Yang et al.~\cite{yang2024mentallama}'s work.
The evaluation was conducted by two experts in the field of psychology. 
The human evaluations are conducted from the following four aspects: 1) \textbf{Consistency:} This dimension assesses whether the explanation is logically coherent and substantiates the final classification decision. 2) \textbf{Reliability:} This dimension assesses whether the content of the explanation is credible, grounded in accurate facts, and supported by sound reasoning. 3) \textbf{Professionality:} This dimension primarily assesses the psychological accuracy and rationality of the generated explanations. 
These three aspects were scored on a four-point scale ranging from 0 to 3, with a higher score indicating a more satisfactory sample in that aspect. 
4) \textbf{Overall:} This dimension assesses the overall effectiveness of the generated explanation and is the average score of consistency, reliability, and professionality. 

\subsection{Ablation experiments}
We conducted ablation experiments to evaluate the impact of the two training steps by creating two separate models: 1) one fine-tuned only on the translated IMHI dataset in the first step, referred to as MentalGLM-chat-S1, and 2) one fine-tuned only on the proposed C-IMHI dataset in the second step, referred to as MentalGLM-chat-S2. 
We then compared these models with the final version, MentalGLM-chat, across three downstream tasks.

\subsection{Downstream task evaluation}
We compared the performance of the MentalGLM series with four representative deep learning models, three generalized LLMs, and two task-specific fine-tuned LLMs across three downstream tasks.
We use the precision, recall and F1-score as the evaluation metrics. 
Note that, this validation only verifies whether it can determine the category, without examining the explanation provided for the output.
The comparison models are introduced below.
\begin{itemize}
\item \textbf{Pre-trained deep learning models:} BERT~\cite{devlin2018bert} represents a Transformer-based~\cite{vaswani2017attention} pre-trained language model. Chinese-BERT-wwm-ext and RoBERTa-wwm~\cite{cui2021pre} represent BERT models optimized for Chinese, which employ whole word masking technology to enhance the ability to understand the Chinese language. Additionally, we selected the SOTA model, Chinese MentalBERT~\cite{zhai-etal-2024-chinese} that enhances text representation and classification capabilities in mental health analysis tasks through continuous pre-training on an extensive corpus of Chinese mental health-related texts.

\item \textbf{Generalized LLMs:} 
LLMs have garnered significant attention due to their flexibility, as they are driven by prompts. Models that perform tasks without examples are referred to as zero-shot prompts, while those that include a few examples are called few-shot prompts. We conducted experiments with three types of generalized LLMs: GLM-4-plus, specifically designed for Chinese, and two from the GPT series, GPT-3.5 and GPT-4. Details of the experimental setup can be found in the works of Qi et al.~\cite{qi2023supervised} and Jiang et al.~\cite{jiang2024ai}.

\item \textbf{Task fine-tuned LLMs:} 
The generalized LLM trained on general corpus without fine-tuning on the target tasks. To ensure fairness, we selected two representative open-source Chinese LLMs: Llama-3-Chinese \footnote{https://huggingface.co/hfl/llama-3-chinese-8b} and Llama-3-Chinese-instruct \footnote{https://huggingface.co/hfl/llama-3-chinese-8b-instruct}~\cite{cui2023efficient}, to perform instruction fine-tuning on the C-IMHI dataset and evaluate them on the downstream tasks. These two models are derived from the existing LLM, Llama3~\cite{dubey2024llama} and have undergone additional fine-tuning and training to better handle Chinese language.

Note that all fine-tuned LLMs were trained on the three downstream tasks simultaneously, whereas the pre-trained deep learning models were trained separately for each task. This also highlights the flexibility of LLMs.

\end{itemize}

\subsection{Quality evaluation of
explanations generated by MentalGLM-chat}
The proposed MentalGLM is capable of generating explanations alongside predictions. For expert evaluation, we randomly selected 100 explanations from MentalGLM-chat, proportionally distributed across the three tasks in the C-IMHI test set: 14 from SOS-HL-1K, 36 from SocialCD-3K, and 50 from the CP dataset.
We evaluated the explanation quality using the same criteria described in Section~\ref{sec:exp:dataset_evaluate:human}, including consistency, reliability and professionality, and averaged these scores for an overall evaluation for each sample.

\subsection{Clinical data evaluation}
Our models were developed with social media data, and we aim to validate their applicability in clinical settings. Thus, we designed this experiment to assess their performance on the cognitive pathway extraction task using clinical data. Due to privacy concerns, online LLMs like GPT-4 cannot be used as baselines.
Therefore, we have selected the following three advanced open-source LLMs for comparison: GLM-4-chat, Baichuan2-Chat \footnote{https://huggingface.co/baichuan-inc/Baichuan2-7B-Chat} ~\cite{yang2023baichuan}, and Qwen2-Instruct \footnote{https://huggingface.co/Qwen/Qwen2-7B-Instruct}~\cite{yang2024qwen2}. 
In addition, we included the task fine-tuned LLM: Llama-3-Chinese-instruct, which was fine-tuned on the C-IMHI dataset and subsequently evaluated on the clinical dataset.
All these comparison models employing the zero-shot prompting strategy, and we reported the performance as micro F1-scores.

\section{Results}

\subsection{Quality evaluation results on the C-IMHI dataset}
\subsubsection{Automatic evaluation} 
\begin{figure*}[h] 
    \centering
    \begin{subfigure}[b]{0.33\textwidth}
        \centering
        \includegraphics[width=\linewidth]{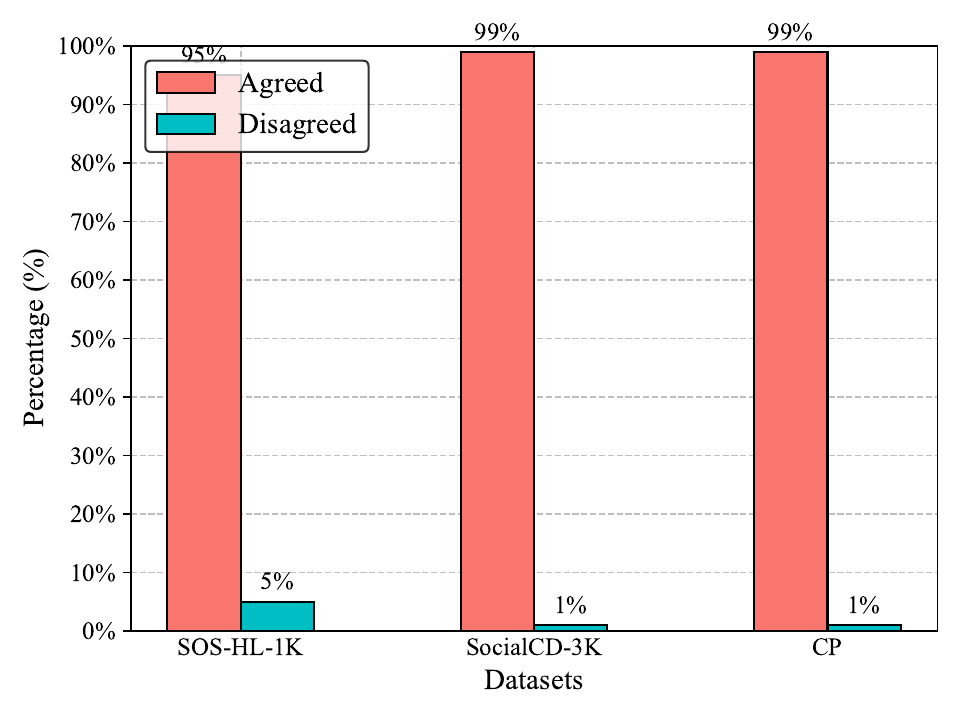} 
        \caption{Correctness} 
        \label{fig:cimhi_correctness}
    \end{subfigure}%
    \hfill 
    \begin{subfigure}[b]{0.33\textwidth}
        \centering
        \includegraphics[width=\linewidth]{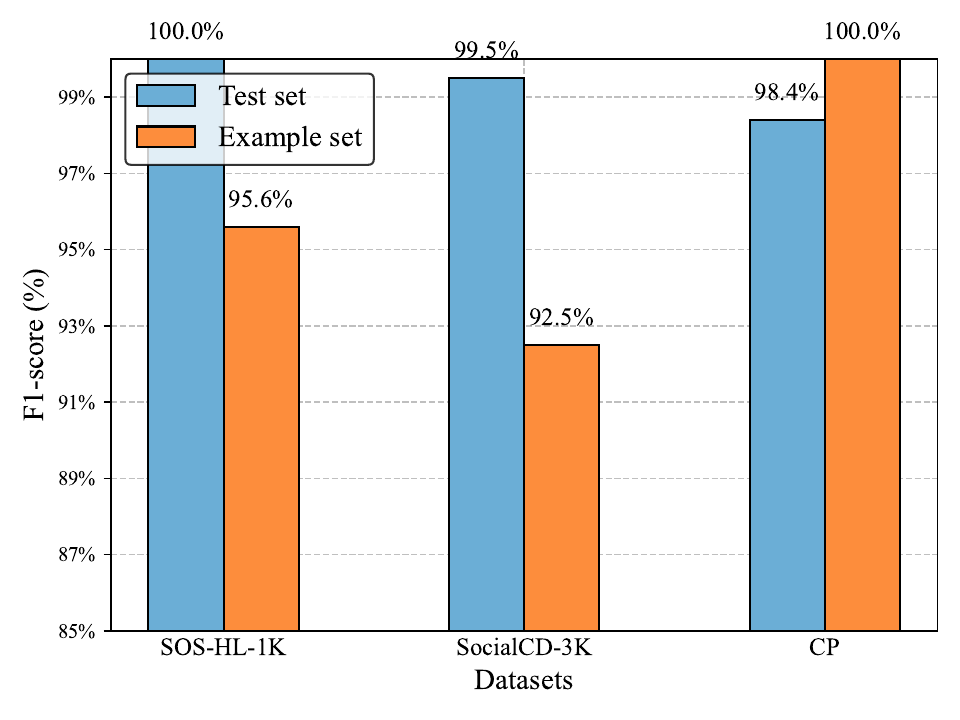} 
        \caption{Consistency} 
        \label{fig:cimhi_consistency}
    \end{subfigure}
    \hfill 
    \begin{subfigure}[b]{0.33\textwidth}
        \centering
        \includegraphics[width=\linewidth]{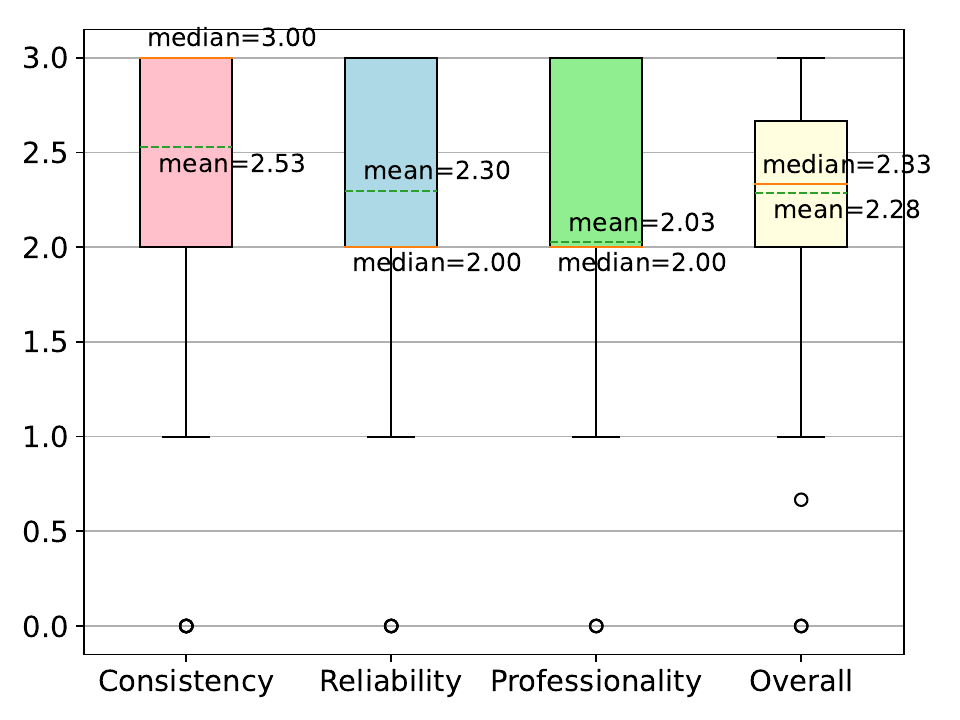} 
        \caption{Human evaluation} 
        \label{fig:cimhi_human_eval}
    \end{subfigure}
    \caption{Evaluation results on GPT-4’s generated explanations used for C-IMHI data construction. (a) and (b) represent automated evaluation for the entire dataset, while (c) shows human evaluation on a subset of 100 samples.}
    \label{fig:cimhi_eval}
\end{figure*}
As described earlier, the proposed dataset C-IMHI was evaluated automatically in terms of correctness and consistency, as shown in Figure~\ref{fig:cimhi_eval}~(a) and~(b), respectively.
We observed that GPT-4 agreed with more than 95\% of the annotations provided by the three datasets, indicating the reliability of the GPT-generated output. However, it is crucial to also assess consistency, ensuring that GPT-4 generates reasoning process explanations that align with both the content and annotations. From the results shown in Figure~\ref{fig:cimhi_eval}~(b), the model achieved more than 98\% F1-score on the test set across the three datasets, indicating high consistency. As we mentioned earlier, this is only a method to estimate consistency, assessing whether the model can identify stable patterns within the dataset, rather than a direct reflection of its performance on downstream tasks. 
The high performance (over 92.5\% F1-score) on the expert-provided example set further confirmed the consistency of our dataset.

\subsubsection{Human evaluation} 
The human evaluation results for C-IMHI dataset are shown in the Figure~\ref{fig:cimhi_eval}~(c).
The high consistency score (mean value >2.5 out of 3) supports the consistency findings obtained from the automatic evaluation.
Additionally, the generated explanations are reliable, as reflected by a high average score of 2.3 out of 3. However, they lack some degree of professionality, as indicated by the slightly lower average score of 2.03.
The overall final score is positive, with a mean of 2.28, indicating that the overall quality of the data has been successfully verified.

\subsection{Ablation experiment results}
The performances of ablation experiments can be seen in the Table \ref{tab:result:ablation}. 
The results reflected the impact of each training stage. The MentalGLM-chat-S1 model, which was only trained with the translated IMHI dataset in the first stage, showed a significant performance gap compared to other models, especially in the cognitive distortion classification task using the SocialCD-3K dataset. This demonstrates the gap between English and Chinese social media contexts, indicating that directly using translated data for fine-tuning does not yield strong performance on downstream tasks.
It's no surprise that fine-tuning and evaluation on the same type of tasks lead to better performance. For instance, MentalGLM-chat-S2 showed a significant improvement compared to MentalGLM-chat-S1, as it benefited from being fine-tuned on data specifically aligned with the target tasks.
However, this does not mean the first step in our method lacks value. Comparing MentalGLM-chat-S2, which was fine-tuned only on C-IMHI, with MentalGLM-chat, which underwent two-step fine-tuning (IMHI and C-IMHI), shows that the first step added valuable knowledge. This improvement is evident in the F1-score increases across the three datasets.

\begin{table}[h]
\centering
\caption{Results of the ablation experiments. All results are reported as F1-scores (\%). MentalGLM-chat-S1 and MentalGLM-chat-S2 refer to the models fine-tuned on the translated IMHI dataset (Stage 1) and the C-IMHI dataset (Stage 2), respectively. ``$CP_{Parent}$'' and ``$CP_{Child}$'' represent parent and child level performance for the $CP$ dataset.}
\resizebox{\columnwidth}{!}{
\begin{tabular}{lcccccc}
\toprule
Model & SOS-HL-1K & SocialCD-3K & $CP_{Parent}$ & $CP_{Child}$ \\
\midrule
MentalGLM-chat-S1 & 66.28 & 15.70 & 48.95 & 21.58 \\
MentalGLM-chat-S2 & 81.58 & 70.69 & 77.69 & 47.69 \\
MentalGLM-chat & 85.12 & 71.04 & 80.55 & 47.85 \\
\bottomrule
\end{tabular}
}
\label{tab:result:ablation}
\end{table}

\begin{table*}[!h]
\caption{Results on the three downstream tasks. We report precision (P), recall (R), and F1-score (F1) as micro averages (in \%), except for the SOS-HL-1K dataset, which uses binary averages. ``$CP_{Parent}$'' and ``$CP_{Child}$'' denote parent and child level performance for the CP dataset. ``ZS'' and ``FS'' represent zero-shot and few shot prompts, respectively.}
\label{tab:result:downstream}
\centering
\scriptsize
\resizebox{\textwidth}{!}{
\begin{tabular}{cccccccccccccc} 
\hline
Method                   & Param & \multicolumn{3}{c}{SOS-HL-1K}                    & \multicolumn{3}{c}{SocialCD-3K}                  & \multicolumn{3}{c}{$CP_{Parent}$}                & \multicolumn{3}{c}{$CP_{Child}$}                  \\ 
\hline
                         &       & F1             & P              & R              & F1             & P              & R              & F1             & P              & R              & F1             & P              & R               \\ 
\hline
\multicolumn{14}{c}{Pre-trained deep learning models}                                                                                                                                                                                         \\ 
\hline
BERT                     & 110M  & 77.91          & 77.60          & 78.22          & 71.92          & 79.19          & 65.87          & 77.01          & 76.68          & 78.13          & 44.66          & 51.84          & 41.85           \\ 
\hline
Chinese-BERT-wwm         & 110M  & 79.32          & 83.18          & 75.80          & 73.06          & \textbf{84.22} & 64.50          & 77.96          & 80.36          & 75.71          & 42.98          & \textbf{71.26} & 30.77           \\ 
\hline
RoBERTa-wwm              & 110M  & 79.83          & 81.51          & 78.22          & 73.91          & 82.61          & 66.87          & 76.81          & 78.28          & 75.39          & 43.61          & 54.81          & 36.21           \\ 
\hline
Chinese MentalBERT       & 110M  & 81.30          & 81.96          & 80.64          & \textbf{74.91} & 83.03          & 68.24          & 79.22          & 79.68          & 78.76          & 47.58          & 50.70          & 44.82           \\ 
\hline
\multicolumn{14}{c}{Generalized LLMs (Zero-shot/few-shot prompt)}                                                                                                                                                                                       \\ 
\hline
GLM-4-plus\_ZS           & -     & 67.23          & 51.50          & \textbf{96.77} & 34.24          & 29.09          & 41.59          & 49.93          & 46.01          & 54.57          & 24.54          & 18.84          & 35.18           \\ 
\hline
GLM-4-plus\_FS           & -     & 68.66          & 54.50          & 92.74          & 41.00          & 32.62          & 55.17          & 43.85          & 44.06          & 43.64          & 27.37          & 21.44          & 37.85           \\ 
\hline
ChatGPT\_ZS              & 175B  & 65.42          & 52.00          & 88.23          & 12.06          & 10.63          & 13.95          & 32.08          & 28.61          & 36.50          & 13.42          & 12.19          & 14.92           \\ 
\hline
ChatGPT\_FS              & 175B  & 68.71          & 59.37          & 81.61          & 18.10          & 16.76          & 19.68          & 54.87          & 51.87          & 58.25          & 27.00          & 24.80          & 29.64           \\ 
\hline
GPT-4\_ZS                & -     & 71.72          & 57.43          & 95.48          & 26.61          & 17.13          & 59.65          & 35.93          & 31.69          & 41.47          & 17.22          & 15.44          & 19.45           \\ 
\hline
GPT-4\_FS                & -     & 75.81          & 70.16          & 82.58          & 40.39          & 31.38          & 56.66          & 59.09          & 55.49          & 63.19          & 28.61          & 23.23          & 37.23           \\ 
\hline
\multicolumn{14}{c}{Task fine-tuned LLMs}                                                                                                                                                                                              \\ 
\hline
Llama-3-Chinese          & 8B    & 77.97          & 82.14          & 74.19          & 70.97          & 74.52          & 67.75          & 79.11          & 79.37          & 78.86          & 48.23          & 48.99          & 47.49           \\ 
\hline
Llama-3-Chinese-instruct & 8B    & 79.32          & 83.19          & 75.81          & 70.90          & 73.05          & 68.87          & 78.96          & 78.42          & 79.50          & 49.32          & 50.16          & 48.51           \\ 
\hline
\multicolumn{14}{c}{Our method}                                                                                                                                                                                                               \\ 
\hline
MentalGLM                & 9B    & 79.20          & 78.57          & 79.84          & 73.06          & 76.71          & \textbf{69.74} & 79.41          & 79.75          & 79.07          & \textbf{50.91} & 51.69          & \textbf{50.15}  \\ 
\hline
MentalGLM-chat           & 9B    & \textbf{85.12} & \textbf{87.29} & 83.06          & 71.04          & 74.22          & 68.12          & \textbf{80.55} & \textbf{80.76} & \textbf{80.34} & 47.85          & 48.42          & 47.28           \\
\hline
\end{tabular}
}
\end{table*}

\subsection{Downstream task results}
The experimental results for the downstream tasks are presented in Table~\ref{tab:result:downstream}.
In general, the performance of our proposed models is accurate, achieving either the best or comparable results across all tasks. For example, in the SOS-HL-1K dataset, our MentalGLM-chat model achieved the highest performance, with an F1-score 3.82\% higher than the SOTA deep learning model, Chinese MentalBERT, which was specifically trained for this task. It also outperformed other generalized LLMs and instruction fine-tuned LLMs. For the SocialCD-3K dataset, our MentalGLM performed similarly to Chinese MentalBERT, with only a 1.85\% point lower F1-score.
For the cognitive pathway extraction task, we evaluated model performance at both parent and child levels. Child-level classification was significantly more challenging, resulting in lower performance. Our model achieved an F1-score of 80.55\% on parent nodes, slightly surpassing the SOTA supervised model. Generalized LLMs, including GPT-4, performed poorly, with the best score at 59.09\%. At the child level, all models struggled, but our MentalGLM outperformed others, including Chinese MentalBERT, by 3.33\%.
In summary, our models outperformed or matched supervised models while providing the flexibility to train multiple tasks simultaneously without separate models. They also outperformed fine-tuned open-source LLMs on all tasks. Although supervised models have fewer parameters, LLMs offer flexibility and explicability by generating decision explanations, a crucial advantage for specific applications as described in the following sections.

\subsection{Evaluation results of explanations generated by MentalGLM-chat}
\label{sec:results:dataset_eval}
\begin{figure}[h]
  \centering
  \includegraphics[width=0.8\linewidth]{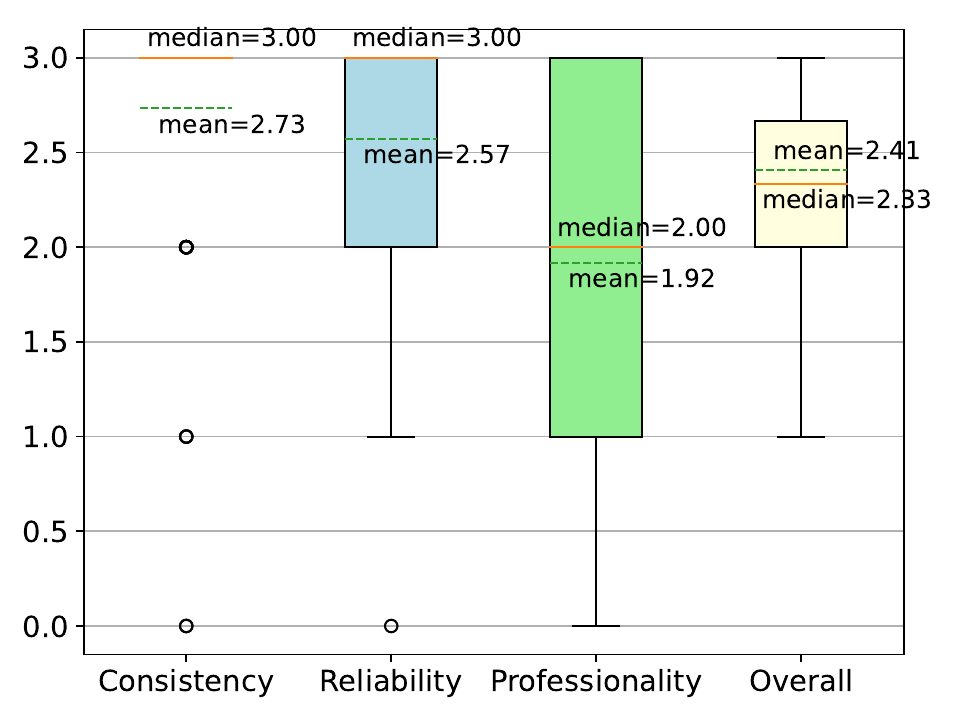}
  \caption{Expert evaluation results on 100 randomly selected prediction explanations generated by MentalGLM-chat.}
  \label{fig:explanation_human_eval}
\end{figure}

As previously mentioned, our model provides not only accurate performance but also predictions with decision-making explanations, crucial for real-world applications.
Figure~\ref{fig:explanation_human_eval} presents the expert evaluation results of 100 randomly selected explanations generated by MentalGLM-chat.
The median values for both consistency and reliability in our expert evaluations were 3, with mean values exceeding 2.5. Compared to the expert evaluations of GPT-4-generated explanations (as shown in Figure~\ref{fig:cimhi_eval}~(c)), MentalGLM outperforms GPT-4 in these two dimensions. However, our model demonstrated lower professionality compared to GPT-4’s explanations. Overall, our model produced better explanations than GPT-4, as indicated by a higher mean value for overall performance.

\subsection{Clinical data evaluation results}
The experimental results of LLMs applied on clinical dataset can be seen in Table~\ref{tab:result:clinical}.
Performance on clinical data for cognitive pathway extraction decreased compared to the CP dataset, but MentalGLM still outperformed other models. Compared to the best performing open-source LLMs without fine-tuning, MentalGLM achieved a 23.79\% increase in F1-score at the parent level and 29.36\% at the child level. Privacy concerns prevented evaluating GPT-4, but results are expected to align with trends. MentalGLM-chat outperformed Llama-3-Chinese-instruct by 4.94\% (parent level) and 2.95\% (child level). Despite lower overall performance compared to public data, MentalGLM demonstrated strong generalisation, showing promise for future use in clinical detection tasks.

\begin{table}[h]
    \centering
    \caption{Results for clinical cognitive pathway extraction tasks . We report the micro F1-score. ``Parent'' and ``Child'' denote parent and child level performance for the clinical dataset.}
    \begin{tabular}{lcccc}
        \toprule
        Model & param & Parent & Child  \\
        \midrule
        Baichuan2-chat & 7B & 41.33 & 12.06  \\
        Qwen2-Instruct & 7B & 44.33 &  11.54 \\
        GLM-4-chat & 9B & 45.28 & 14.97  \\
        Llama-3-Chinese-instruct & 8B & 64.25 & 42.41  \\
        MentalGLM & 9B & 69.07& 44.33 \\
        MentalGLM-chat & 9B & 69.19 & 45.36  \\
        \bottomrule
    \end{tabular}
    \label{tab:result:clinical}
\end{table}

\section{Discussion}

In this paper, we introduce C-IMHI, a manually evaluated dataset for instruction fine-tuning in mental health analysis, and the MentalGLM series models, which achieved strong performance across three tasks, with F1-scores of 85.12\% on SOS-HL-1K, 73.06\% on SocialCD-3K, and 80.55\% and 50.91\% on CP-parent and CP-child, respectively. Our models outperformed or matched other deep learning models and provided decision-making explanations. To our knowledge, no other dataset exists for Chinese interpretable mental health analysis from social media, and we are the first to develop task-specific, fine-tuned LLMs for this purpose. We also evaluated the model on a clinical dataset for cognitive pathway extraction, where it achieved the best performance, demonstrating strong generalizability.

Compared to deep learning methods that treat the task as classification, we frame it as a generative task, predicting both categories and explanations for greater flexibility. Domain-specific LLMs adapt to multiple tasks, unlike supervised learning which requires separate models, limiting generalisation. The generated explanations scored well in consistency (2.73/3) and reliability (2.57/3) but lacked professionality (1.92/3). To improve, we will incorporate more expert feedback during fine-tuning and explore reinforcement learning from human feedback (RLHF). Despite outperforming other models on clinical data, a gap remains compared to public datasets, which we plan to address using domain transfer techniques.

\section{Conclusion}
In summary, we introduced C-IMHI, the first dataset of interpretable mental health analysis instructions specifically focused on Chinese social media. Then based on this dataset, we introduced MentalGLM, the first open-source series of LLMs designed for explainable mental health analysis on Chinese social media. The experimental results indicated that MentalGLM achieved better or comparable performance compared to SOTA discriminant deep learning methods and other LLMs, while also generating high-quality explanations. Furthermore, we validated the model's performance on clinical datasets, where its strong performance highlighted its generalizability, suggesting potential applicability for clinical use. All the experimental datasets and proposed models are publicly available to support future research.

\section{Acknowledgments}\label{sec:acknowledgments}
This work was supported by grants from the National Natural Science Foundation of China (grant numbers: 72304212 and 82071546), Fundamental Research Funds for the Central Universities (grant numbers: 2042022kf1218; 2042022kf1037), the Young Top-notch Talent Cultivation Program of Hubei Province.
Guanghui Fu is supported by a Chinese Government Scholarship provided by the China Scholarship Council (CSC).

\bibliographystyle{ACM-Reference-Format}
\bibliography{ref}


\begin{thebibliography}{42}


\ifx \showCODEN    \undefined \def \showCODEN     #1{\unskip}     \fi
\ifx \showDOI      \undefined \def \showDOI       #1{#1}\fi
\ifx \showISBNx    \undefined \def \showISBNx     #1{\unskip}     \fi
\ifx \showISBNxiii \undefined \def \showISBNxiii  #1{\unskip}     \fi
\ifx \showISSN     \undefined \def \showISSN      #1{\unskip}     \fi
\ifx \showLCCN     \undefined \def \showLCCN      #1{\unskip}     \fi
\ifx \shownote     \undefined \def \shownote      #1{#1}          \fi
\ifx \showarticletitle \undefined \def \showarticletitle #1{#1}   \fi
\ifx \showURL      \undefined \def \showURL       {\relax}        \fi
\providecommand\bibfield[2]{#2}
\providecommand\bibinfo[2]{#2}
\providecommand\natexlab[1]{#1}
\providecommand\showeprint[2][]{arXiv:#2}

\bibitem[Beck(1970)]%
        {beck1970cognitive}
\bibfield{author}{\bibinfo{person}{Aaron~T Beck}.} \bibinfo{year}{1970}\natexlab{}.
\newblock \showarticletitle{Cognitive therapy: Nature and relation to behavior therapy}.
\newblock \bibinfo{journal}{\emph{Behavior therapy}} \bibinfo{volume}{1}, \bibinfo{number}{2} (\bibinfo{year}{1970}), \bibinfo{pages}{184--200}.
\newblock


\bibitem[Brown(2020)]%
        {brown2020language}
\bibfield{author}{\bibinfo{person}{Tom~B Brown}.} \bibinfo{year}{2020}\natexlab{}.
\newblock \showarticletitle{Language models are few-shot learners}.
\newblock \bibinfo{journal}{\emph{arXiv preprint arXiv:2005.14165}} (\bibinfo{year}{2020}).
\newblock


\bibitem[Burns(1981)]%
        {burns1981feeling}
\bibfield{author}{\bibinfo{person}{David~D Burns}.} \bibinfo{year}{1981}\natexlab{}.
\newblock \bibinfo{booktitle}{\emph{Feeling good}}.
\newblock \bibinfo{publisher}{Signet Book}.
\newblock


\bibitem[Cao et~al\mbox{.}(2019)]%
        {cao2019latent}
\bibfield{author}{\bibinfo{person}{Lei Cao}, \bibinfo{person}{Huijun Zhang}, \bibinfo{person}{Ling Feng}, \bibinfo{person}{Zihan Wei}, \bibinfo{person}{Xin Wang}, \bibinfo{person}{Ningyun Li}, {and} \bibinfo{person}{Xiaohao He}.} \bibinfo{year}{2019}\natexlab{}.
\newblock \showarticletitle{Latent Suicide Risk Detection on Microblog via Suicide-Oriented Word Embeddings and Layered Attention}. In \bibinfo{booktitle}{\emph{Proceedings of the 2019 Conference on Empirical Methods in Natural Language Processing and the 9th International Joint Conference on Natural Language Processing (EMNLP-IJCNLP)}}, \bibfield{editor}{\bibinfo{person}{Kentaro Inui}, \bibinfo{person}{Jing Jiang}, \bibinfo{person}{Vincent Ng}, {and} \bibinfo{person}{Xiaojun Wan}} (Eds.). \bibinfo{publisher}{Association for Computational Linguistics}, \bibinfo{address}{Hong Kong, China}, \bibinfo{pages}{1718--1728}.
\newblock
\urldef\tempurl%
\url{https://doi.org/10.18653/v1/D19-1181}
\showDOI{\tempurl}


\bibitem[Chen et~al\mbox{.}(2023)]%
        {chen2023soulchat}
\bibfield{author}{\bibinfo{person}{Yirong Chen}, \bibinfo{person}{Xiaofen Xing}, \bibinfo{person}{Jingkai Lin}, \bibinfo{person}{Huimin Zheng}, \bibinfo{person}{Zhenyu Wang}, \bibinfo{person}{Qi Liu}, {and} \bibinfo{person}{Xiangmin Xu}.} \bibinfo{year}{2023}\natexlab{}.
\newblock \showarticletitle{{S}oul{C}hat: Improving {LLM}s{'} Empathy, Listening, and Comfort Abilities through Fine-tuning with Multi-turn Empathy Conversations}. In \bibinfo{booktitle}{\emph{Findings of the Association for Computational Linguistics: EMNLP 2023}}, \bibfield{editor}{\bibinfo{person}{Houda Bouamor}, \bibinfo{person}{Juan Pino}, {and} \bibinfo{person}{Kalika Bali}} (Eds.). \bibinfo{publisher}{Association for Computational Linguistics}, \bibinfo{address}{Singapore}, \bibinfo{pages}{1170--1183}.
\newblock
\urldef\tempurl%
\url{https://doi.org/10.18653/v1/2023.findings-emnlp.83}
\showDOI{\tempurl}


\bibitem[Coppersmith et~al\mbox{.}(2018)]%
        {coppersmith2018natural}
\bibfield{author}{\bibinfo{person}{Glen Coppersmith}, \bibinfo{person}{Ryan Leary}, \bibinfo{person}{Patrick Crutchley}, {and} \bibinfo{person}{Alex Fine}.} \bibinfo{year}{2018}\natexlab{}.
\newblock \showarticletitle{Natural language processing of social media as screening for suicide risk}.
\newblock \bibinfo{journal}{\emph{Biomedical informatics insights}}  \bibinfo{volume}{10} (\bibinfo{year}{2018}), \bibinfo{pages}{1178222618792860}.
\newblock


\bibitem[Cui et~al\mbox{.}(2021)]%
        {cui2021pre}
\bibfield{author}{\bibinfo{person}{Yiming Cui}, \bibinfo{person}{Wanxiang Che}, \bibinfo{person}{Ting Liu}, \bibinfo{person}{Bing Qin}, {and} \bibinfo{person}{Ziqing Yang}.} \bibinfo{year}{2021}\natexlab{}.
\newblock \showarticletitle{Pre-training with whole word masking for chinese bert}.
\newblock \bibinfo{journal}{\emph{IEEE/ACM Transactions on Audio, Speech, and Language Processing}}  \bibinfo{volume}{29} (\bibinfo{year}{2021}), \bibinfo{pages}{3504--3514}.
\newblock


\bibitem[Cui et~al\mbox{.}(2023)]%
        {cui2023efficient}
\bibfield{author}{\bibinfo{person}{Yiming Cui}, \bibinfo{person}{Ziqing Yang}, {and} \bibinfo{person}{Xin Yao}.} \bibinfo{year}{2023}\natexlab{}.
\newblock \showarticletitle{Efficient and effective text encoding for chinese llama and alpaca}.
\newblock \bibinfo{journal}{\emph{arXiv preprint arXiv:2304.08177}} (\bibinfo{year}{2023}).
\newblock


\bibitem[Demszky et~al\mbox{.}(2023)]%
        {demszky2023using}
\bibfield{author}{\bibinfo{person}{Dorottya Demszky}, \bibinfo{person}{Diyi Yang}, \bibinfo{person}{David~S Yeager}, \bibinfo{person}{Christopher~J Bryan}, \bibinfo{person}{Margarett Clapper}, \bibinfo{person}{Susannah Chandhok}, \bibinfo{person}{Johannes~C Eichstaedt}, \bibinfo{person}{Cameron Hecht}, \bibinfo{person}{Jeremy Jamieson}, \bibinfo{person}{Meghann Johnson}, {et~al\mbox{.}}} \bibinfo{year}{2023}\natexlab{}.
\newblock \showarticletitle{Using large language models in psychology}.
\newblock \bibinfo{journal}{\emph{Nature Reviews Psychology}} \bibinfo{volume}{2}, \bibinfo{number}{11} (\bibinfo{year}{2023}), \bibinfo{pages}{688--701}.
\newblock


\bibitem[Devlin(2018)]%
        {devlin2018bert}
\bibfield{author}{\bibinfo{person}{Jacob Devlin}.} \bibinfo{year}{2018}\natexlab{}.
\newblock \showarticletitle{Bert: Pre-training of deep bidirectional transformers for language understanding}.
\newblock \bibinfo{journal}{\emph{arXiv preprint arXiv:1810.04805}} (\bibinfo{year}{2018}).
\newblock


\bibitem[Dubey et~al\mbox{.}(2024)]%
        {dubey2024llama}
\bibfield{author}{\bibinfo{person}{Abhimanyu Dubey}, \bibinfo{person}{Abhinav Jauhri}, \bibinfo{person}{Abhinav Pandey}, \bibinfo{person}{Abhishek Kadian}, \bibinfo{person}{Ahmad Al-Dahle}, \bibinfo{person}{Aiesha Letman}, \bibinfo{person}{Akhil Mathur}, \bibinfo{person}{Alan Schelten}, \bibinfo{person}{Amy Yang}, \bibinfo{person}{Angela Fan}, {et~al\mbox{.}}} \bibinfo{year}{2024}\natexlab{}.
\newblock \showarticletitle{The {Llama} 3 herd of models}.
\newblock \bibinfo{journal}{\emph{arXiv preprint arXiv:2407.21783}} (\bibinfo{year}{2024}).
\newblock


\bibitem[Freeman(2022)]%
        {world2022world}
\bibfield{author}{\bibinfo{person}{Melvyn Freeman}.} \bibinfo{year}{2022}\natexlab{}.
\newblock \showarticletitle{The World Mental Health Report: transforming mental health for all}.
\newblock \bibinfo{journal}{\emph{World Psychiatry}} \bibinfo{volume}{21}, \bibinfo{number}{3} (\bibinfo{year}{2022}), \bibinfo{pages}{391--392}.
\newblock
\urldef\tempurl%
\url{https://doi.org/10.1002/wps.21018}
\showDOI{\tempurl}
\showeprint{https://onlinelibrary.wiley.com/doi/pdf/10.1002/wps.21018}


\bibitem[Garg et~al\mbox{.}(2022)]%
        {garg-etal-2022-cams}
\bibfield{author}{\bibinfo{person}{Muskan Garg}, \bibinfo{person}{Chandni Saxena}, \bibinfo{person}{Sriparna Saha}, \bibinfo{person}{Veena Krishnan}, \bibinfo{person}{Ruchi Joshi}, {and} \bibinfo{person}{Vijay Mago}.} \bibinfo{year}{2022}\natexlab{}.
\newblock \showarticletitle{{CAMS}: An Annotated Corpus for Causal Analysis of Mental Health Issues in Social Media Posts}. In \bibinfo{booktitle}{\emph{Proceedings of the Thirteenth Language Resources and Evaluation Conference}}, \bibfield{editor}{\bibinfo{person}{Nicoletta Calzolari}, \bibinfo{person}{Fr{\'e}d{\'e}ric B{\'e}chet}, \bibinfo{person}{Philippe Blache}, \bibinfo{person}{Khalid Choukri}, \bibinfo{person}{Christopher Cieri}, \bibinfo{person}{Thierry Declerck}, \bibinfo{person}{Sara Goggi}, \bibinfo{person}{Hitoshi Isahara}, \bibinfo{person}{Bente Maegaard}, \bibinfo{person}{Joseph Mariani}, \bibinfo{person}{H{\'e}l{\`e}ne Mazo}, \bibinfo{person}{Jan Odijk}, {and} \bibinfo{person}{Stelios Piperidis}} (Eds.). \bibinfo{publisher}{European Language Resources Association}, \bibinfo{address}{Marseille, France}, \bibinfo{pages}{6387--6396}.
\newblock
\urldef\tempurl%
\url{https://aclanthology.org/2022.lrec-1.686}
\showURL{%
\tempurl}


\bibitem[Garg et~al\mbox{.}(2023)]%
        {garg-etal-2023-annotated}
\bibfield{author}{\bibinfo{person}{Muskan Garg}, \bibinfo{person}{Amirmohammad Shahbandegan}, \bibinfo{person}{Amrit Chadha}, {and} \bibinfo{person}{Vijay Mago}.} \bibinfo{year}{2023}\natexlab{}.
\newblock \showarticletitle{An Annotated Dataset for Explainable Interpersonal Risk Factors of Mental Disturbance in Social Media Posts}. In \bibinfo{booktitle}{\emph{Findings of the Association for Computational Linguistics: ACL 2023}}, \bibfield{editor}{\bibinfo{person}{Anna Rogers}, \bibinfo{person}{Jordan Boyd-Graber}, {and} \bibinfo{person}{Naoaki Okazaki}} (Eds.). \bibinfo{publisher}{Association for Computational Linguistics}, \bibinfo{address}{Toronto, Canada}, \bibinfo{pages}{11960--11969}.
\newblock
\urldef\tempurl%
\url{https://doi.org/10.18653/v1/2023.findings-acl.757}
\showDOI{\tempurl}


\bibitem[GLM et~al\mbox{.}(2024)]%
        {glm2024chatglm}
\bibfield{author}{\bibinfo{person}{Team GLM}, \bibinfo{person}{Aohan Zeng}, \bibinfo{person}{Bin Xu}, \bibinfo{person}{Bowen Wang}, \bibinfo{person}{Chenhui Zhang}, \bibinfo{person}{Da Yin}, \bibinfo{person}{Diego Rojas}, \bibinfo{person}{Guanyu Feng}, \bibinfo{person}{Hanlin Zhao}, \bibinfo{person}{Hanyu Lai}, {et~al\mbox{.}}} \bibinfo{year}{2024}\natexlab{}.
\newblock \showarticletitle{Chat{GLM}: A Family of Large Language Models from GLM-130B to GLM-4 All Tools}.
\newblock \bibinfo{journal}{\emph{arXiv preprint arXiv:2406.12793}} (\bibinfo{year}{2024}).
\newblock


\bibitem[He et~al\mbox{.}(2023)]%
        {he2023towards}
\bibfield{author}{\bibinfo{person}{Tianyu He}, \bibinfo{person}{Guanghui Fu}, \bibinfo{person}{Yijing Yu}, \bibinfo{person}{Fan Wang}, \bibinfo{person}{Jianqiang Li}, \bibinfo{person}{Qing Zhao}, \bibinfo{person}{Changwei Song}, \bibinfo{person}{Hongzhi Qi}, \bibinfo{person}{Dan Luo}, \bibinfo{person}{Huijing Zou}, {et~al\mbox{.}}} \bibinfo{year}{2023}\natexlab{}.
\newblock \showarticletitle{Towards a psychological generalist ai: A survey of current applications of large language models and future prospects}.
\newblock \bibinfo{journal}{\emph{arXiv preprint arXiv:2312.04578}} (\bibinfo{year}{2023}).
\newblock


\bibitem[Hu et~al\mbox{.}(2022)]%
        {hu2022lora}
\bibfield{author}{\bibinfo{person}{Edward~J Hu}, \bibinfo{person}{yelong shen}, \bibinfo{person}{Phillip Wallis}, \bibinfo{person}{Zeyuan Allen-Zhu}, \bibinfo{person}{Yuanzhi Li}, \bibinfo{person}{Shean Wang}, \bibinfo{person}{Lu Wang}, {and} \bibinfo{person}{Weizhu Chen}.} \bibinfo{year}{2022}\natexlab{}.
\newblock \showarticletitle{Lo{RA}: Low-Rank Adaptation of Large Language Models}. In \bibinfo{booktitle}{\emph{International Conference on Learning Representations}}.
\newblock
\urldef\tempurl%
\url{https://openreview.net/forum?id=nZeVKeeFYf9}
\showURL{%
\tempurl}


\bibitem[Huang et~al\mbox{.}(2019)]%
        {huang2019prevalence}
\bibfield{author}{\bibinfo{person}{Yueqin Huang}, \bibinfo{person}{YU Wang}, \bibinfo{person}{Hong Wang}, \bibinfo{person}{Zhaorui Liu}, \bibinfo{person}{Xin Yu}, \bibinfo{person}{Jie Yan}, \bibinfo{person}{Yaqin Yu}, \bibinfo{person}{Changgui Kou}, \bibinfo{person}{Xiufeng Xu}, \bibinfo{person}{Jin Lu}, {et~al\mbox{.}}} \bibinfo{year}{2019}\natexlab{}.
\newblock \showarticletitle{Prevalence of mental disorders in China: a cross-sectional epidemiological study}.
\newblock \bibinfo{journal}{\emph{The Lancet Psychiatry}} \bibinfo{volume}{6}, \bibinfo{number}{3} (\bibinfo{year}{2019}), \bibinfo{pages}{211--224}.
\newblock


\bibitem[Ji et~al\mbox{.}(2022a)]%
        {ji2022suicidal}
\bibfield{author}{\bibinfo{person}{Shaoxiong Ji}, \bibinfo{person}{Xue Li}, \bibinfo{person}{Zi Huang}, {and} \bibinfo{person}{Erik Cambria}.} \bibinfo{year}{2022}\natexlab{a}.
\newblock \showarticletitle{Suicidal ideation and mental disorder detection with attentive relation networks}.
\newblock \bibinfo{journal}{\emph{Neural Computing and Applications}} \bibinfo{volume}{34}, \bibinfo{number}{13} (\bibinfo{year}{2022}), \bibinfo{pages}{10309--10319}.
\newblock


\bibitem[Ji et~al\mbox{.}(2022b)]%
        {ji2021mentalbert}
\bibfield{author}{\bibinfo{person}{Shaoxiong Ji}, \bibinfo{person}{Tianlin Zhang}, \bibinfo{person}{Luna Ansari}, \bibinfo{person}{Jie Fu}, \bibinfo{person}{Prayag Tiwari}, {and} \bibinfo{person}{Erik Cambria}.} \bibinfo{year}{2022}\natexlab{b}.
\newblock \showarticletitle{{M}ental{BERT}: Publicly Available Pretrained Language Models for Mental Healthcare}. In \bibinfo{booktitle}{\emph{Proceedings of the Thirteenth Language Resources and Evaluation Conference}}, \bibfield{editor}{\bibinfo{person}{Nicoletta Calzolari}, \bibinfo{person}{Fr{\'e}d{\'e}ric B{\'e}chet}, \bibinfo{person}{Philippe Blache}, \bibinfo{person}{Khalid Choukri}, \bibinfo{person}{Christopher Cieri}, \bibinfo{person}{Thierry Declerck}, \bibinfo{person}{Sara Goggi}, \bibinfo{person}{Hitoshi Isahara}, \bibinfo{person}{Bente Maegaard}, \bibinfo{person}{Joseph Mariani}, \bibinfo{person}{H{\'e}l{\`e}ne Mazo}, \bibinfo{person}{Jan Odijk}, {and} \bibinfo{person}{Stelios Piperidis}} (Eds.). \bibinfo{publisher}{European Language Resources Association}, \bibinfo{address}{Marseille, France}, \bibinfo{pages}{7184--7190}.
\newblock
\urldef\tempurl%
\url{https://aclanthology.org/2022.lrec-1.778}
\showURL{%
\tempurl}


\bibitem[Jiang et~al\mbox{.}(2024)]%
        {jiang2024ai}
\bibfield{author}{\bibinfo{person}{Meng Jiang}, \bibinfo{person}{Yi~Jing Yu}, \bibinfo{person}{Qing Zhao}, \bibinfo{person}{Jianqiang Li}, \bibinfo{person}{Changwei Song}, \bibinfo{person}{Hongzhi Qi}, \bibinfo{person}{Wei Zhai}, \bibinfo{person}{Dan Luo}, \bibinfo{person}{Xiaoqin Wang}, \bibinfo{person}{Guanghui Fu}, {et~al\mbox{.}}} \bibinfo{year}{2024}\natexlab{}.
\newblock \showarticletitle{{AI}-Enhanced Cognitive Behavioral Therapy: Deep Learning and Large Language Models for Extracting Cognitive Pathways from Social Media Texts}.
\newblock \bibinfo{journal}{\emph{arXiv preprint arXiv:2404.11449}} (\bibinfo{year}{2024}).
\newblock


\bibitem[Loshchilov and Hutter(2019)]%
        {loshchilov2018decoupled}
\bibfield{author}{\bibinfo{person}{Ilya Loshchilov} {and} \bibinfo{person}{Frank Hutter}.} \bibinfo{year}{2019}\natexlab{}.
\newblock \showarticletitle{Decoupled Weight Decay Regularization}. In \bibinfo{booktitle}{\emph{International Conference on Learning Representations}}.
\newblock
\urldef\tempurl%
\url{https://openreview.net/forum?id=Bkg6RiCqY7}
\showURL{%
\tempurl}


\bibitem[Mauriello et~al\mbox{.}(2021)]%
        {mauriello2021sad}
\bibfield{author}{\bibinfo{person}{Matthew~Louis Mauriello}, \bibinfo{person}{Thierry Lincoln}, \bibinfo{person}{Grace Hon}, \bibinfo{person}{Dorien Simon}, \bibinfo{person}{Dan Jurafsky}, {and} \bibinfo{person}{Pablo Paredes}.} \bibinfo{year}{2021}\natexlab{}.
\newblock \showarticletitle{{SAD}: A Stress Annotated Dataset for Recognizing Everyday Stressors in SMS-like Conversational Systems}. In \bibinfo{booktitle}{\emph{Extended Abstracts of the 2021 CHI Conference on Human Factors in Computing Systems}} (Yokohama, Japan) \emph{(\bibinfo{series}{CHI EA '21})}. \bibinfo{publisher}{Association for Computing Machinery}, \bibinfo{address}{New York, NY, USA}, Article \bibinfo{articleno}{399}, \bibinfo{numpages}{7}~pages.
\newblock
\showISBNx{9781450380959}
\urldef\tempurl%
\url{https://doi.org/10.1145/3411763.3451799}
\showDOI{\tempurl}


\bibitem[Organization et~al\mbox{.}(2023)]%
        {world2023depressive}
\bibfield{author}{\bibinfo{person}{World~Health Organization}, \bibinfo{person}{World~Health Organization}, {et~al\mbox{.}}} \bibinfo{year}{2023}\natexlab{}.
\newblock \showarticletitle{Depressive disorder (depression). 2023}.
\newblock \bibinfo{journal}{\emph{Retrieved from Depressive disorder (depression)(who. int)}} (\bibinfo{year}{2023}).
\newblock


\bibitem[Pirina and {\c{C}}{\"o}ltekin(2018)]%
        {pirina2018identifying}
\bibfield{author}{\bibinfo{person}{Inna Pirina} {and} \bibinfo{person}{{\c{C}}a{\u{g}}r{\i} {\c{C}}{\"o}ltekin}.} \bibinfo{year}{2018}\natexlab{}.
\newblock \showarticletitle{Identifying Depression on {R}eddit: The Effect of Training Data}. In \bibinfo{booktitle}{\emph{Proceedings of the 2018 {EMNLP} Workshop {SMM}4{H}: The 3rd Social Media Mining for Health Applications Workshop {\&} Shared Task}}, \bibfield{editor}{\bibinfo{person}{Graciela Gonzalez-Hernandez}, \bibinfo{person}{Davy Weissenbacher}, \bibinfo{person}{Abeed Sarker}, {and} \bibinfo{person}{Michael Paul}} (Eds.). \bibinfo{publisher}{Association for Computational Linguistics}, \bibinfo{address}{Brussels, Belgium}, \bibinfo{pages}{9--12}.
\newblock
\urldef\tempurl%
\url{https://doi.org/10.18653/v1/W18-5903}
\showDOI{\tempurl}


\bibitem[Qi et~al\mbox{.}(2023)]%
        {qi2023supervised}
\bibfield{author}{\bibinfo{person}{Hongzhi Qi}, \bibinfo{person}{Qing Zhao}, \bibinfo{person}{Jianqiang Li}, \bibinfo{person}{Changwei Song}, \bibinfo{person}{Wei Zhai}, \bibinfo{person}{Luo Dan}, \bibinfo{person}{Shuo Liu}, \bibinfo{person}{Yi~Jing Yu}, \bibinfo{person}{Fan Wang}, \bibinfo{person}{Huijing Zou}, {et~al\mbox{.}}} \bibinfo{year}{2023}\natexlab{}.
\newblock \showarticletitle{Supervised Learning and Large Language Model Benchmarks on Mental Health Datasets: Cognitive Distortions and Suicidal Risks in Chinese Social Media}.
\newblock  (\bibinfo{year}{2023}).
\newblock


\bibitem[Sathvik and Garg(2023)]%
        {sathvik2023multiwd}
\bibfield{author}{\bibinfo{person}{MSVPJ Sathvik} {and} \bibinfo{person}{Muskan Garg}.} \bibinfo{year}{2023}\natexlab{}.
\newblock \showarticletitle{Multiwd: Multiple wellness dimensions in social media posts}.
\newblock \bibinfo{journal}{\emph{Authorea Preprints}} (\bibinfo{year}{2023}).
\newblock


\bibitem[Sheu(2020)]%
        {sheu2020illuminating}
\bibfield{author}{\bibinfo{person}{Yi-han Sheu}.} \bibinfo{year}{2020}\natexlab{}.
\newblock \showarticletitle{Illuminating the black box: interpreting deep neural network models for psychiatric research}.
\newblock \bibinfo{journal}{\emph{Frontiers in Psychiatry}}  \bibinfo{volume}{11} (\bibinfo{year}{2020}), \bibinfo{pages}{551299}.
\newblock


\bibitem[Turcan and McKeown(2019)]%
        {turcan2019dreaddit}
\bibfield{author}{\bibinfo{person}{Elsbeth Turcan} {and} \bibinfo{person}{Kathy McKeown}.} \bibinfo{year}{2019}\natexlab{}.
\newblock \showarticletitle{{D}readdit: A {R}eddit Dataset for Stress Analysis in Social Media}. In \bibinfo{booktitle}{\emph{Proceedings of the Tenth International Workshop on Health Text Mining and Information Analysis (LOUHI 2019)}}, \bibfield{editor}{\bibinfo{person}{Eben Holderness}, \bibinfo{person}{Antonio Jimeno~Yepes}, \bibinfo{person}{Alberto Lavelli}, \bibinfo{person}{Anne-Lyse Minard}, \bibinfo{person}{James Pustejovsky}, {and} \bibinfo{person}{Fabio Rinaldi}} (Eds.). \bibinfo{publisher}{Association for Computational Linguistics}, \bibinfo{address}{Hong Kong}, \bibinfo{pages}{97--107}.
\newblock
\urldef\tempurl%
\url{https://doi.org/10.18653/v1/D19-6213}
\showDOI{\tempurl}


\bibitem[Vaswani(2017)]%
        {vaswani2017attention}
\bibfield{author}{\bibinfo{person}{A Vaswani}.} \bibinfo{year}{2017}\natexlab{}.
\newblock \showarticletitle{Attention is all you need}.
\newblock \bibinfo{journal}{\emph{Advances in Neural Information Processing Systems}} (\bibinfo{year}{2017}).
\newblock


\bibitem[Wang et~al\mbox{.}(2023)]%
        {wang2023selfconsistency}
\bibfield{author}{\bibinfo{person}{Xuezhi Wang}, \bibinfo{person}{Jason Wei}, \bibinfo{person}{Dale Schuurmans}, \bibinfo{person}{Quoc~V Le}, \bibinfo{person}{Ed~H. Chi}, \bibinfo{person}{Sharan Narang}, \bibinfo{person}{Aakanksha Chowdhery}, {and} \bibinfo{person}{Denny Zhou}.} \bibinfo{year}{2023}\natexlab{}.
\newblock \showarticletitle{Self-Consistency Improves Chain of Thought Reasoning in Language Models}. In \bibinfo{booktitle}{\emph{The Eleventh International Conference on Learning Representations}}.
\newblock
\urldef\tempurl%
\url{https://openreview.net/forum?id=1PL1NIMMrw}
\showURL{%
\tempurl}


\bibitem[Wei et~al\mbox{.}(2022)]%
        {wei2022finetuned}
\bibfield{author}{\bibinfo{person}{Jason Wei}, \bibinfo{person}{Maarten Bosma}, \bibinfo{person}{Vincent Zhao}, \bibinfo{person}{Kelvin Guu}, \bibinfo{person}{Adams~Wei Yu}, \bibinfo{person}{Brian Lester}, \bibinfo{person}{Nan Du}, \bibinfo{person}{Andrew~M. Dai}, {and} \bibinfo{person}{Quoc~V Le}.} \bibinfo{year}{2022}\natexlab{}.
\newblock \showarticletitle{Finetuned Language Models are Zero-Shot Learners}. In \bibinfo{booktitle}{\emph{International Conference on Learning Representations}}.
\newblock
\urldef\tempurl%
\url{https://openreview.net/forum?id=gEZrGCozdqR}
\showURL{%
\tempurl}


\bibitem[Xu et~al\mbox{.}(2024)]%
        {xu2024mental}
\bibfield{author}{\bibinfo{person}{Xuhai Xu}, \bibinfo{person}{Bingsheng Yao}, \bibinfo{person}{Yuanzhe Dong}, \bibinfo{person}{Saadia Gabriel}, \bibinfo{person}{Hong Yu}, \bibinfo{person}{James Hendler}, \bibinfo{person}{Marzyeh Ghassemi}, \bibinfo{person}{Anind~K Dey}, {and} \bibinfo{person}{Dakuo Wang}.} \bibinfo{year}{2024}\natexlab{}.
\newblock \showarticletitle{{Mental-LLM}: Leveraging large language models for mental health prediction via online text data}.
\newblock \bibinfo{journal}{\emph{Proceedings of the ACM on Interactive, Mobile, Wearable and Ubiquitous Technologies}} \bibinfo{volume}{8}, \bibinfo{number}{1} (\bibinfo{year}{2024}), \bibinfo{pages}{1--32}.
\newblock


\bibitem[Yang et~al\mbox{.}(2023b)]%
        {yang2023baichuan}
\bibfield{author}{\bibinfo{person}{Aiyuan Yang}, \bibinfo{person}{Bin Xiao}, \bibinfo{person}{Bingning Wang}, \bibinfo{person}{Borong Zhang}, \bibinfo{person}{Ce Bian}, \bibinfo{person}{Chao Yin}, \bibinfo{person}{Chenxu Lv}, \bibinfo{person}{Da Pan}, \bibinfo{person}{Dian Wang}, \bibinfo{person}{Dong Yan}, {et~al\mbox{.}}} \bibinfo{year}{2023}\natexlab{b}.
\newblock \showarticletitle{Baichuan 2: Open large-scale language models}.
\newblock \bibinfo{journal}{\emph{arXiv preprint arXiv:2309.10305}} (\bibinfo{year}{2023}).
\newblock


\bibitem[Yang et~al\mbox{.}(2024a)]%
        {yang2024qwen2}
\bibfield{author}{\bibinfo{person}{An Yang}, \bibinfo{person}{Baosong Yang}, \bibinfo{person}{Binyuan Hui}, \bibinfo{person}{Bo Zheng}, \bibinfo{person}{Bowen Yu}, \bibinfo{person}{Chang Zhou}, \bibinfo{person}{Chengpeng Li}, \bibinfo{person}{Chengyuan Li}, \bibinfo{person}{Dayiheng Liu}, \bibinfo{person}{Fei Huang}, {et~al\mbox{.}}} \bibinfo{year}{2024}\natexlab{a}.
\newblock \showarticletitle{Qwen2 technical report}.
\newblock \bibinfo{journal}{\emph{arXiv preprint arXiv:2407.10671}} (\bibinfo{year}{2024}).
\newblock


\bibitem[Yang et~al\mbox{.}(2023a)]%
        {yang-etal-2023-towards}
\bibfield{author}{\bibinfo{person}{Kailai Yang}, \bibinfo{person}{Shaoxiong Ji}, \bibinfo{person}{Tianlin Zhang}, \bibinfo{person}{Qianqian Xie}, \bibinfo{person}{Ziyan Kuang}, {and} \bibinfo{person}{Sophia Ananiadou}.} \bibinfo{year}{2023}\natexlab{a}.
\newblock \showarticletitle{Towards Interpretable Mental Health Analysis with Large Language Models}. In \bibinfo{booktitle}{\emph{Proceedings of the 2023 Conference on Empirical Methods in Natural Language Processing}}, \bibfield{editor}{\bibinfo{person}{Houda Bouamor}, \bibinfo{person}{Juan Pino}, {and} \bibinfo{person}{Kalika Bali}} (Eds.). \bibinfo{publisher}{Association for Computational Linguistics}, \bibinfo{address}{Singapore}, \bibinfo{pages}{6056--6077}.
\newblock
\urldef\tempurl%
\url{https://doi.org/10.18653/v1/2023.emnlp-main.370}
\showDOI{\tempurl}


\bibitem[Yang et~al\mbox{.}(2024b)]%
        {yang2024mentallama}
\bibfield{author}{\bibinfo{person}{Kailai Yang}, \bibinfo{person}{Tianlin Zhang}, \bibinfo{person}{Ziyan Kuang}, \bibinfo{person}{Qianqian Xie}, \bibinfo{person}{Jimin Huang}, {and} \bibinfo{person}{Sophia Ananiadou}.} \bibinfo{year}{2024}\natexlab{b}.
\newblock \showarticletitle{{MentaLLaMA}: Interpretable Mental Health Analysis on Social Media with Large Language Models}. In \bibinfo{booktitle}{\emph{Proceedings of the ACM Web Conference 2024}} (Singapore, Singapore) \emph{(\bibinfo{series}{WWW '24})}. \bibinfo{publisher}{Association for Computing Machinery}, \bibinfo{address}{New York, NY, USA}, \bibinfo{pages}{4489–4500}.
\newblock
\showISBNx{9798400701719}
\urldef\tempurl%
\url{https://doi.org/10.1145/3589334.3648137}
\showDOI{\tempurl}


\bibitem[Yang et~al\mbox{.}(2024c)]%
        {yang2024zhongjing}
\bibfield{author}{\bibinfo{person}{Songhua Yang}, \bibinfo{person}{Hanjie Zhao}, \bibinfo{person}{Senbin Zhu}, \bibinfo{person}{Guangyu Zhou}, \bibinfo{person}{Hongfei Xu}, \bibinfo{person}{Yuxiang Jia}, {and} \bibinfo{person}{Hongying Zan}.} \bibinfo{year}{2024}\natexlab{c}.
\newblock \showarticletitle{Zhongjing: Enhancing the Chinese Medical Capabilities of Large Language Model through Expert Feedback and Real-World Multi-Turn Dialogue}.
\newblock \bibinfo{journal}{\emph{Proceedings of the AAAI Conference on Artificial Intelligence}} \bibinfo{volume}{38}, \bibinfo{number}{17} (\bibinfo{date}{Mar.} \bibinfo{year}{2024}), \bibinfo{pages}{19368--19376}.
\newblock
\urldef\tempurl%
\url{https://doi.org/10.1609/aaai.v38i17.29907}
\showDOI{\tempurl}


\bibitem[Yu et~al\mbox{.}(2020)]%
        {yu2020coping}
\bibfield{author}{\bibinfo{person}{Hua Yu}, \bibinfo{person}{Mingli Li}, \bibinfo{person}{Zhixiong Li}, \bibinfo{person}{Weiyi Xiang}, \bibinfo{person}{Yiwen Yuan}, \bibinfo{person}{Yaya Liu}, \bibinfo{person}{Zhe Li}, {and} \bibinfo{person}{Zhenzhen Xiong}.} \bibinfo{year}{2020}\natexlab{}.
\newblock \showarticletitle{Coping style, social support and psychological distress in the general Chinese population in the early stages of the {COVID-19} epidemic}.
\newblock \bibinfo{journal}{\emph{BMC psychiatry}}  \bibinfo{volume}{20} (\bibinfo{year}{2020}), \bibinfo{pages}{1--11}.
\newblock


\bibitem[Yu et~al\mbox{.}(2024)]%
        {yu2024large}
\bibfield{author}{\bibinfo{person}{Yue Yu}, \bibinfo{person}{Yuchen Zhuang}, \bibinfo{person}{Jieyu Zhang}, \bibinfo{person}{Yu Meng}, \bibinfo{person}{Alexander Ratner}, \bibinfo{person}{Ranjay Krishna}, \bibinfo{person}{Jiaming Shen}, {and} \bibinfo{person}{Chao Zhang}.} \bibinfo{year}{2024}\natexlab{}.
\newblock \showarticletitle{Large language model as attributed training data generator: a tale of diversity and bias}. In \bibinfo{booktitle}{\emph{Proceedings of the 37th International Conference on Neural Information Processing Systems}} (New Orleans, LA, USA) \emph{(\bibinfo{series}{NIPS '23})}. \bibinfo{publisher}{Curran Associates Inc.}, \bibinfo{address}{Red Hook, NY, USA}, Article \bibinfo{articleno}{2433}, \bibinfo{numpages}{51}~pages.
\newblock


\bibitem[Zhai et~al\mbox{.}(2024)]%
        {zhai-etal-2024-chinese}
\bibfield{author}{\bibinfo{person}{Wei Zhai}, \bibinfo{person}{Hongzhi Qi}, \bibinfo{person}{Qing Zhao}, \bibinfo{person}{Jianqiang Li}, \bibinfo{person}{Ziqi Wang}, \bibinfo{person}{Han Wang}, \bibinfo{person}{Bing Yang}, {and} \bibinfo{person}{Guanghui Fu}.} \bibinfo{year}{2024}\natexlab{}.
\newblock \showarticletitle{{C}hinese {M}ental{BERT}: Domain-Adaptive Pre-training on Social Media for {C}hinese Mental Health Text Analysis}. In \bibinfo{booktitle}{\emph{Findings of the Association for Computational Linguistics ACL 2024}}, \bibfield{editor}{\bibinfo{person}{Lun-Wei Ku}, \bibinfo{person}{Andre Martins}, {and} \bibinfo{person}{Vivek Srikumar}} (Eds.). \bibinfo{publisher}{Association for Computational Linguistics}, \bibinfo{address}{Bangkok, Thailand and virtual meeting}, \bibinfo{pages}{10574--10585}.
\newblock
\urldef\tempurl%
\url{https://aclanthology.org/2024.findings-acl.629}
\showURL{%
\tempurl}


\bibitem[Zhou et~al\mbox{.}(2023)]%
        {zhou2024lima}
\bibfield{author}{\bibinfo{person}{Chunting Zhou}, \bibinfo{person}{Pengfei Liu}, \bibinfo{person}{Puxin Xu}, \bibinfo{person}{Srinivasan Iyer}, \bibinfo{person}{Jiao Sun}, \bibinfo{person}{Yuning Mao}, \bibinfo{person}{Xuezhe Ma}, \bibinfo{person}{Avia Efrat}, \bibinfo{person}{Ping Yu}, \bibinfo{person}{LILI YU}, \bibinfo{person}{Susan Zhang}, \bibinfo{person}{Gargi Ghosh}, \bibinfo{person}{Mike Lewis}, \bibinfo{person}{Luke Zettlemoyer}, {and} \bibinfo{person}{Omer Levy}.} \bibinfo{year}{2023}\natexlab{}.
\newblock \showarticletitle{LIMA: Less Is More for Alignment}. In \bibinfo{booktitle}{\emph{Advances in Neural Information Processing Systems}}, \bibfield{editor}{\bibinfo{person}{A.~Oh}, \bibinfo{person}{T.~Naumann}, \bibinfo{person}{A.~Globerson}, \bibinfo{person}{K.~Saenko}, \bibinfo{person}{M.~Hardt}, {and} \bibinfo{person}{S.~Levine}} (Eds.), Vol.~\bibinfo{volume}{36}. \bibinfo{publisher}{Curran Associates, Inc.}, \bibinfo{pages}{55006--55021}.
\newblock
\urldef\tempurl%
\url{https://proceedings.neurips.cc/paper_files/paper/2023/file/ac662d74829e4407ce1d126477f4a03a-Paper-Conference.pdf}
\showURL{%
\tempurl}


\end{thebibliography}
\newpage
\appendix

\renewcommand\thefigure{S\arabic{figure}} 
\setcounter{figure}{0} \renewcommand\thetable{S\arabic{table}}
\renewcommand\thesection{Appendix \Alph{section}} 

\setcounter{table}{0}

\section{Ethical considerations}
The original datasets used to construct the C-IMHI dataset were sourced from public social media platforms. We adhere strictly to privacy protocols and ethical principles to protect user privacy. To minimize the risk of personal information leakage, we anonymize and de-identify the data extensively during processing and analysis. We ensure that the research findings do not include any information that can directly or indirectly identify an individual. For the clinical data, informed consent was obtained from both hospitals and patients, and all data were anonymized to safeguard patient privacy.

Although MentalGLM has shown promising results in both social media and clinical mental health tasks, it is important to acknowledge that LLMs may introduce potential biases, including those related to gender, age, or sexual orientation, which could lead to incorrect judgments and inappropriate interpretations. We emphasize that the use of experimental results and data is strictly confined to research and analysis purposes, and any misuse or improper handling of the information is explicitly prohibited.

\section{Prompt template for explanation generation using GPT-4} \label{sec:append:template}
\begin{figure}[h]
  \centering
  \includegraphics[width=0.8\linewidth]{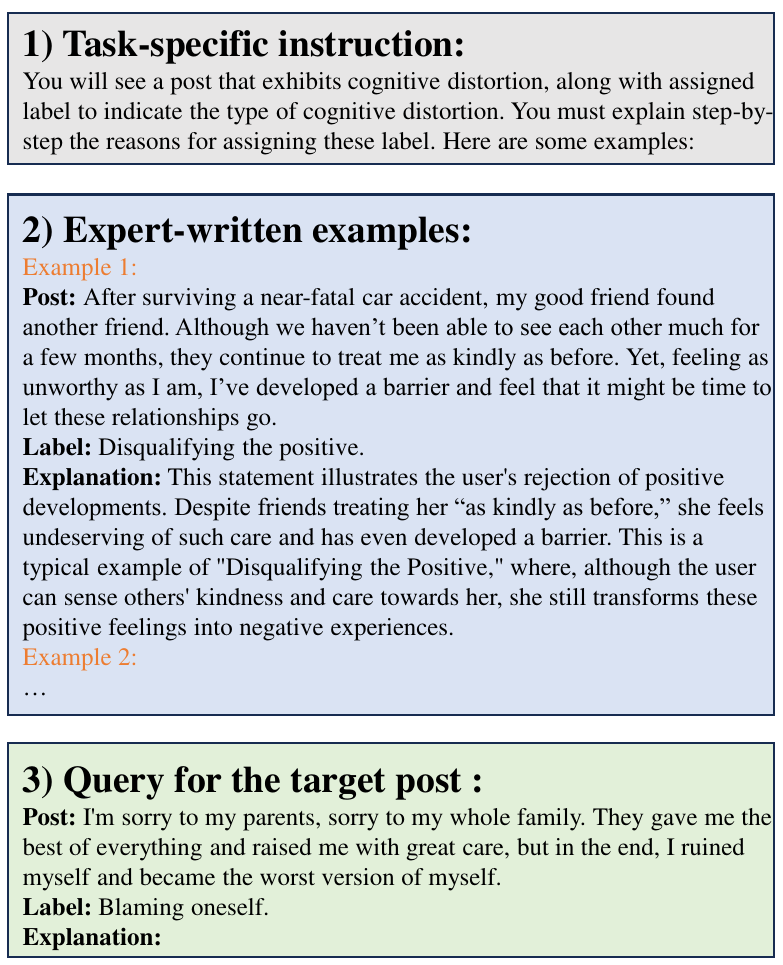}
  \caption{Prompt template used in GPT-4 to generate explanations for cognitive distortion task.}
  \label{fig:appendix:explain_template}
\end{figure}

The prompt structure consists of the following three parts: 1) Task-specific instruction: This defines the task. 2) Expert-written examples: These enable GPT-4 to learn and imitate the thinking of experts for this task. 3) Query for the target post: This specifies the samples that need to be analyzed and explained, including post text and corresponding labels. Figure~\ref{fig:appendix:explain_template} illustrates the prompt template employed by GPT-4 to generate explanations for cognitive distortion task.

\section{Consistency evaluation of the C-IMHI dataset} \label{sec:append:quality_eval}

We proposed a method to check the consistency of the generated explanations by verifying whether each explanation supports its respective label. To achieve this, we trained deep learning models using explanation-label pairs to predict the corresponding labels from the generated explanations. The model used for this task was Chinese MentalBERT. The performance on the test set reflects the consistency of the generated explanations.
The model that performed best on the validation set was selected and evaluated on both the test set and the expert-provided example set. We conducted this evaluation on three downstream task datasets: SOS-HL-1K, SocialCD-3K, and CP, maintaining the dataset distribution as shown in Table~\ref{tab:dataset_chinese}.

For training, the models were fine-tuned for 10 epochs on the training set for all tasks. We used a batch size of 16, the Adam optimizer, a learning rate of 2e-5, and cross-entropy as the loss function. Training was conducted on an NVIDIA GeForce RTX 4090 24GB GPU. The best-performing model on the validation set was then used to evaluate the test set and the expert-provided example set, with F1-scores used as the evaluation metric.

We can see the detailed model performance in Table~\ref{tab:appendix:consistency_eval}. 
The high performance (>98\% F1-score) across all the test sets demonstrates the strong consistency of the generated explanations. Additionally, the high performance (>92\% F1-score) on the expert-provided example set highlights the consistency of the model's explanations with human-provided explanations.
\begin{table}[!htbp]
\caption{Performance of the Chinese MentalBERT classifier on the test and expert-provided example sets, evaluated using precision (P), recall (R), and F1-score (F1). Metrics are reported as micro averages, except for binary averages on SOS-HL-1K.}
\label{tab:appendix:consistency_eval}
\centering
\resizebox{\linewidth}{!}{%
\begin{tabular}{ccccccccc} 
\hline
\multicolumn{3}{c}{\textbf{SOS-HL-1K}} & \multicolumn{3}{c}{\textbf{SocialCD-3K}} & \multicolumn{3}{c}{\textbf{CP}}        \\ 
\hline
\textbf{F1} & \textbf{P} & \textbf{R}  & \textbf{F1} & \textbf{P} & \textbf{R}    & \textbf{F1} & \textbf{P} & \textbf{R}  \\ 
\hline
\multicolumn{9}{c}{Test set}                                                                                               \\ 
\hline
100.00      & 100.00     & 100.00      & 99.56       & 99.25      & 99.87         & 98.47       & 99.81      & 97.16       \\ 
\hline
\multicolumn{9}{c}{Expert-provided example set}                                                                            \\ 
\hline
95.65       & 91.66      & 100.00      & 92.53       & 93.93      & 91.17         & 100.00      & 100.00     & 100.00      \\
\hline
\end{tabular}
}
\end{table}

\end{document}